\documentclass[11pt]{article}

\usepackage[final]{acl}

\usepackage{times}
\usepackage{latexsym}
\usepackage[T1]{fontenc}

\usepackage[utf8]{inputenc}

\usepackage{microtype}

\usepackage{inconsolata}

\usepackage{graphicx}

\usepackage[normalem]{ulem}
\usepackage{xurl}
\usepackage{placeins}
\usepackage{xcolor}
\usepackage{soul}
\sethlcolor{yellow}

\usepackage{xcolor,color,xspace,enumerate,centernot,multirow,float,graphicx,
xcolor,caption,subcaption,textcomp,pgfplots,pgf-pie,tikz,listings,
comment,adjustbox,mdframed,changepage,algorithm,algorithmic}
\usepackage{tikz}
\usetikzlibrary{arrows.meta, positioning}
\usepackage{msc}
\usepackage{enumitem}
\usepackage{listings}

\usepackage{tablefootnote}
\usepackage{verbatim}
\usepackage{booktabs}
\definecolor{gris}{gray}{0.85}

\newcommand{\var}[1]{\textbf{\texttt{<#1>}}}
\usepackage[most]{tcolorbox}
\newtcolorbox{promptbox}{
  colback=gray!5,
  colframe=gray!40,
  boxrule=0.5pt,
  arc=2mm,
  left=3mm,
  right=3mm,
  top=1mm,
  bottom=1mm,
  fontupper=\small,
}

\title{RFCLLM: Evaluating  LLMs’ Reasoning Ability of Network Protocol State Machines}

\author{
Anqi Chen$^{1}$ \qquad
Dan Goldwasser$^{2}$ \qquad
Cristina Nita-Rotaru$^{1}$ \\
Northeastern University$^{1}$ \qquad
Purdue University$^{2}$ \\
\texttt{\{chen.anqi3, c.nitarotaru\}@northeastern.edu}
\qquad
\texttt{dgoldwas@purdue.edu}
}

\begin{document}
\maketitle

\begin{abstract}
    Mapping textual specifications into formal representations is essential for ensuring the correctness of protocol designs and implementations. 
     LLM-generated mappings, used for networking security or testing, are assumed to capture a perfect understanding of the specification, which may not hold in practice. The goal of this paper is to assess the extent to which LLMs can interpret the specification correctly. We examine the degree to which an LLM’s implicit representation of a finite-state transition system–defined via natural language descriptions–aligns with a manually generated ground-truth model. 

    We designed 4 tasks and 1482 task queries for 16 protocols. We evaluated different judge biases, observed the inherent difficulty gaps between tasks, looked into the effect of 4 context types, and the influence of protocol characteristics. Our work contributes to a step toward verifying whether LLMs can really be trusted in FSM (Finite State Machine) reasoning of protocol specifications.
\end{abstract}
\section{Introduction}

Natural language is the most common way to describe network protocol specifications because it is easily understood by different domain experts. These specifications consist of prose documents augmented with pseudocode, header descriptions, or state machines. However, textual specifications cannot be directly used to verify protocol correctness or determine whether an implementation conforms to the specification. Such analysis requires formal verification tools that operate on precise formal languages and grammars. Thus, mapping textual specifications into formal representations is essential for ensuring the correctness of protocol designs and implementations.

In the past, mapping protocol specifications to formal models was performed manually \cite{snake_dsn2015,tcpwn_ndss2018} or using NLP techniques \cite{rfcnlp_sp2022}. In recent years, LLMs have been increasingly used to generate such mappings that support formal verification, testing, and security, or are directly integrated into these workflows \cite{chatafl, rfcaudit, formalllm, llmmodel}. Unlike earlier NLP approaches that were carefully evaluated for their ability to interpret protocol specifications, existing work that uses LLM outputs for networking security or testing often assumes LLMs have a perfect understanding of the specifications and can correctly map high-level technical descriptions into a formal model: this may not hold in practice. 
This paper assesses the extent to which LLMs can interpret the specification correctly. We focus on Request for Comments (RFCs), the most well-known specifications for Internet Protocols. 

\textbf{Motivating Example.} To help clarify the inherent ambiguity of textual specification and how domain knowledge helps clarify the author's intent, consider the following example from \cite{sctp_2024} identifying an ambiguity in RFC 9260 (SCTP) when describing how a peer should react upon receiving an unexpected INIT message:
{{\fontfamily{ppl}\selectfont
\textit{
Upon receipt of an INIT chunk in the Cookie\_Echoed state, an endpoint MUST respond with an INIT\_ACK chunk using \underline{the same parameters} it sent in its original INIT chunk.
}
}}To provide relevant context, SCTP uses two security identifiers, i-tag (Initiate Tag) and v-tag (Verification Tag), to establish and validate an active connection between two endpoints, by using the i-tag sent by one peer as the v-tag of the other, and vice-versa. 

\begin{figure}
  \begin{center}
\includegraphics[width=0.35\textwidth]{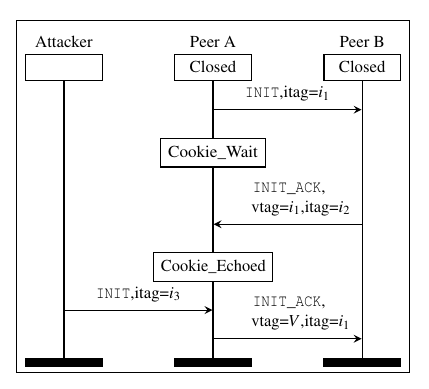}
  \end{center}
  \vspace{-15pt}
    \caption{SCTP Ambiguity \cite{sctp_2024}.}
    \label{fig:sctp}
      \vspace{-5pt}
\end{figure}

Fig. \ref{fig:sctp} shows two peers, A and B, initially both in Closed state, and an attacker who can spoof the port and IP of B. 
At the end of the sequence, what value should the v-tag V take?  The phrase "same parameters" is ambiguous: A reasonable reader might set V = i1; the correct answer is V = i2, but this is clear only if the reader understands that i-tags and v-tags serve opposite roles across the two directions of the connection. An implementer who follows the incorrect interpretation allows an attacker to inject forged packets and hijack the connection.

Our proposed evaluation directly targets such scenarios, systematically evaluating the legality and completeness of transitions afforded by the LLM interpretation of the text.



\textbf{Problem Formulation.}
%
We frame the challenge as an assessment of world model comprehension \cite{vafa2024evaluating}. Each RFC defines a world model in natural language, which can be represented formally as a finite-state transition system, consisting of states, variables, and transition logic. We propose four types of QA tasks over state transitions and their conditions (e.g., \textit{given two states, what is the sequence of events that leads from one state to the other?)}. These tasks serve as a starting point towards an evaluation of LLMs' understanding of network protocol semantics, and are directly applicable to networking security tasks, such as protocol fuzzing exploration \cite{chatafl, llmiot, statepre, llmassistedmodelbasedfuzzingprotocol}, trace analysis \cite{llmdreamsockets}, or formal analysis \cite{llmmodel, llmtemporal}. 

To create the QA data, we collected 16 protocols covering different layers in the Open Systems Interconnection network stack, resulting in a total sum of 118 states, 334 state transitions, 5381 valid paths, and 1482 task queries. We annotated each protocol with a ground-truth FSM to derive the gold answers. When comparing the generated responses to the gold answers, we employed two types of automated judges, measuring the cosine similarity between the two~\cite{PSMBench}, and framing it as a paraphrase detection task over the gold and model-generated answers pair.

\textbf{Modeling Approach.} Our analysis is designed to assess the degree to which an LLM’s implicit representation of the world model after reading the RFC description aligns with the manually generated ground-truth.  Unlike prior research that induces world models through labeled training data \cite{vafa2024evaluating} or reasoning over concise graph definitions \cite{wang2023can, zhang-etal-2024-llm-graph, yuan-etal-2025-gracore}, 
our work focuses on realistic settings that reflect the ambiguity and underspecified nature of technical documentation. Because RFCs define high-level behaviors and rely on human domain knowledge to resolve inconsistencies, we examine whether LLMs can bridge these interpretive gaps.
To this end, we evaluate several strategies for eliciting these implicit protocol models. A baseline that relies on the model’s parameters (\textit{No-context}) without additional information, and a skyline using an idealized, unambiguous transition table (\textit{Easy}). We compare these against two document-grounded representations: one where pertinent RFC sections are manually curated by experts (\textit{Manual-extracted-RFC}) and another where they are automatically retrieved by an LLM (\textit{LLM-extracted-RFC}). 

Our experimental settings directly aim to separate between memorisation, retrieval, and reasoning. The \textit{No-context} setting captures memorisation (since no textual specification is added to the prompt, models have to rely on the protocol knowledge captured through their parameters). Comparing \textit{Manual-extracted-RFC} and \textit{LLM-extracted-RFC} examines the difference between perfect (expert-curated) and realistic (automatically-extracted) retrieval. Finally, reasoning is evaluated by looking at the difference between \textit{Easy} and \textit{Manual-extracted-RFC}, which captures the difference between reasoning over the gold RFC model (i.e., logical reasoning) or over the perfectly retrieved textual description of the model (i.e., textual reasoning). Comparing these settings allows us to distinguish memorization from specification-based reasoning, and assess whether failures stem from a lack of structural reasoning or from an inability to navigate the dense prose of technical standards.


%

     \textbf{Experimental Study.} We evaluated 8 LLMs, including two open-source and six commercial models covering different vendors and parameter sizes.
     Our experimental analysis is centered around five questions: (1) \textit{How do different judge methods behave?} (2) \textit{How does the difficulty of different tasks vary? (3) How do different context types affect the performance? }(4)\textit{ How do different protocols' characteristics affect the performance?} (5) \textit{And how do different LLMs perform in answering the tasks?}

Our evaluation compared the bias of different judges and the inherent difficulty gaps between tasks, highlighting three main findings: (1) document understanding is the primary bottleneck, as models perform substantially better with structured FSMs (\emph{Easy} context) than raw RFC prose; (2) manually curated RFC excerpts consistently outperform automatically extracted contexts, demonstrating that information extraction remains challenging; and (3) multi-hop reasoning remains difficult even when the correct FSM (\emph{Easy} context) is provided. Among all protocols, specific per-protocol characteristics play an important role in performance, and we identified some key factors involved. 


\textbf{Key contributions.} We suggest an evaluation framework for LLMs' reasoning abilities in 
network protocol FSM comprehension, consisting of 4 tasks of varying difficulty. We suggest four context settings, designed to highlight different aspects of models' reasoning challenges. Our work resulted in a comprehensive dataset, consisting of 1482 task queries derived from 16 protocols. Our analysis revealed insights about the varying abilities of 8 models in FSM understanding.
%
%

We believe these insights translate to many language comprehension scenarios that require models to follow a strict textual specification, such as following game instructions and constrained decoding applications. We released our code at: \url{https://anonymous.4open.science/r/rfcllm}.



\section{Related Work}
We explicitly separate two lines of research: (1) work on extracting or constructing protocol FSMs from specification text, and (2) work that directly analyzes or reasons over protocol specifications for security applications. We further position our work as the first benchmark that evaluates protocol reasoning capabilities rather than FSM construction.

For the first category, recent work explores the use of large language models (LLMs) for specification understanding and model extraction. For example, PROSPER \cite{prosper_hotnet2023} shows how LLMs can help extract structured protocol specifications from RFCs despite ambiguity and distributed descriptions of protocol behavior across the document. PSMBench \cite{PSMBench} provides a dataset pairing annotated RFC text with manually extracted ground truth protocol state machines, enabling the evaluation of LLMs' performance on automatic FSM extraction.

For the second category, NLP techniques including LLMs are employed to analyse specifications for security applications. RFCNLP \cite{rfcnlp_sp2022} is a hybrid approach that extracts FSMs from RFC documents using a combination of word representations, zero-shot learning, and rule-based mapping from extracted information into a protocol FSM. Their system demonstrates that FSMs extracted from RFCs can be used to automatically synthesize attacks against protocols such as TCP and DCCP. RFCAUDIT \cite{rfcaudit} checks the implementation code against RFC definitions in an automatic agentic way to identify implementation discrepancies with specifications and potential resulting bugs.

Previous work evaluates whether LLMs can construct protocol FSMs from RFC text. Our benchmark addresses a fundamentally different question: once a protocol specification (or even the correct FSM itself) is available, can LLMs correctly reason about the protocol semantics encoded in it? This distinction is motivated by how LLMs are commonly used in networking practice. Many recent systems directly provide RFC text or protocol specifications to an LLM and implicitly assume that the model can correctly understand the protocol semantics. Our results show that even under the \textit{Easy} setting, where the correct FSM is explicitly provided, models still struggle with multi-hop path reasoning, revealing a gap between constructing protocol representations and reasoning over protocol semantics.

    

    

\section{Approach}

\begin{figure}[htbp]
    \centering
    \includegraphics[width=\linewidth]{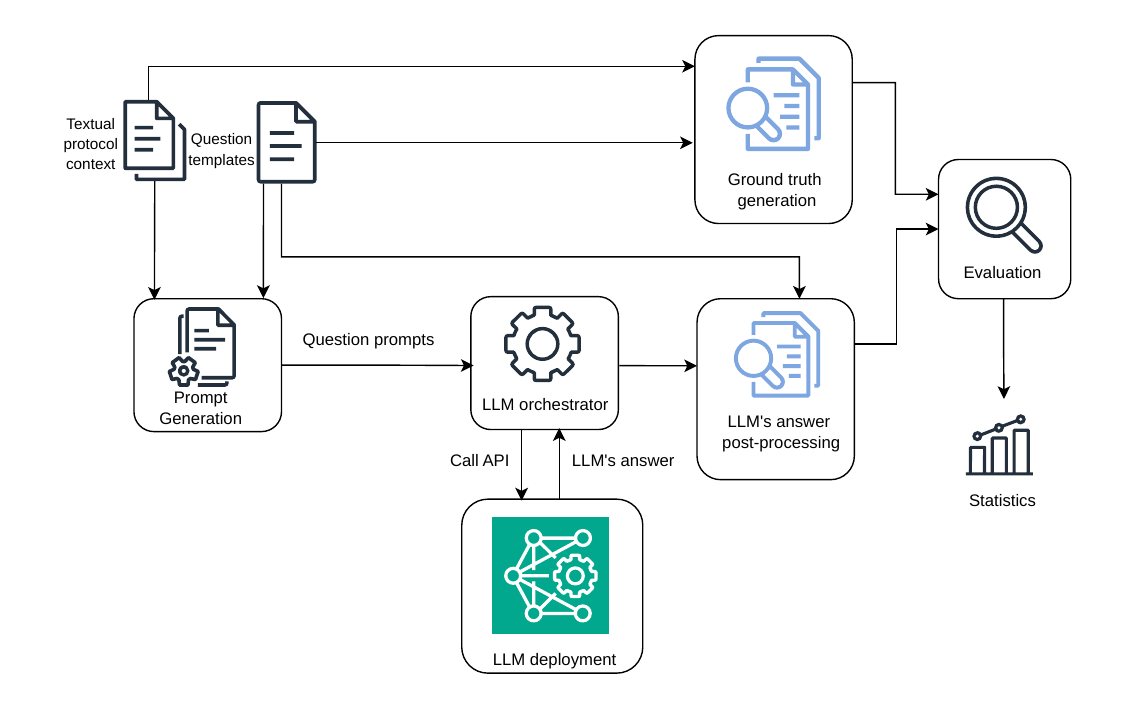}
    \caption{RFCLLM Workflow}
    \label{fig:workflow}
\end{figure}

To measure the LLM's understanding of network protocol FSM logic, we define several tasks and compare the different context information options. Finally, we compare the results of the LLM query with the ground truth and analyze the statistics. Figure \ref{fig:workflow} shows the workflow of our approach.

\paragraph{Ground truth FSM:} 
We define an FSM as a set of transition tuples: i.e. 
$\text{FSM}=\{\, \langle s,\gamma,s',A' \rangle 
\;|\;s, s'\in S,\, \gamma \in C,\,\, A'\subseteq A \,\}$

where: $S$ is the set of states; $C$ is the set of transition conditions; $A$ is the set of actions (e.g. updating variable values, sending packets), and $A' \subseteq A$ is the set of actions triggered by the transition;  $\langle s, \gamma, s', A' \rangle$ is a transition tuple: from the current state $s$, under condition $\gamma$, the FSM moves to the next state $s'$ and performs actions in $A'$.


For example, TCP Congestion Control Protocol's FSM, one transition tuple $\langle s, \gamma, s', A' \rangle$, can be \text{$s$:} Exponential Backoff;
\text{$\gamma$:} ACK; \text{$s'$:} Slow Start;
\text{$A’$:} [Reset rto timer, Send new packets as allowed].

Our gold FSMs were constructed by one author with four years of Ph.D. research experience in networking and security, together with a formal undergraduate education in computer science. The construction procedure depended on how each protocol was specified: when an RFC contained an explicit ASCII FSM, we directly extracted its states and transitions; when the FSM was represented as transition tables or other structured descriptions, we manually converted those representations into our transition-tuple format; for the TCP congestion control protocol, we extracted transition rules from previously published protocol analysis papers. For protocols whose state logic was distributed across RFC prose, we used the PSMBench \cite{PSMBench} FSMs as an initial reference, carefully verified them against the corresponding RFCs, and corrected errors and omissions that we identified. 


 
 

\paragraph{Contexts provided to LLM:} When evaluating LLM's capabilities, different helpful contexts are provided to the LLM  besides the task prompt:

    $\bullet$ \textbf{Easy}: manually crafted state transition table.
    
    $\bullet$ \textbf{Manual-extracted-RFC}: manually extracted textual description from the RFC document.
    
     $\bullet$ \textbf{LLM-extracted-RFC}: LLM-extracted textual description from the uploaded RFC document; here, we used ChatGPT5 for the purpose.
    
    $\bullet$ \textbf{No-context}: no context at all, LLMs answer only according to their own knowledge.

\paragraph{Tasks:} To measure LLMs' understanding, we design several tasks, expressed by question templates and instantiated for each protocol using its FSM.
We include the template prompt details in Appendix \ref{app:prompt}. The tasks require the model to reason about valid transitions (task 1),  the conditions enabling these transitions (task 2), backward transitions, i.e., ``how we got here'' (task 3), and identify all valid paths between two states (task 4). These tasks abstract over the reasoning steps used by common networking security applications, which often employ LLMs.  

\noindent\textbf{Task 1 (Next-state-action Prediction)}: For a specific protocol, given the current state $s$ and a condition $\gamma$, ask the LLM to predict both the next state $s'$ the protocol should transition into and the resulting actions $A'$ caused by the transition.

\noindent\textbf{Task 2 (Condition Inference)}: For a specific protocol, given the current state $s$ and a next state $s'$ the protocol transitions into; ask the LLM what possible condition(s) $C'$ can cause that.

\noindent\textbf{Task 3 (Prev.-state Inference)}: For a specific protocol, given the transition condition $c$ and the resulting state $s'$ the protocol transitions into, ask the LLM what possible previous state(s) $S_{i}$ it was in.

\noindent\textbf{Task 4 (Paths Prediction)}: For a specific protocol, given a current state $s$ and a future state $s'$ the protocol transitions into; ask the LLM what are the possible simple paths from $s$ to $s'$. A simple path is a sequence:
$\langle s$; condition1; state1; condition2; state2; ...; $s' \rangle$ where no loop exists.

 \paragraph{Judges:} We compare the LLM's answers with the ground truth information by using judges. We use (sim-0.6, Llama3.3-70b) for tasks 1, 2, 3, and (Llama3.3-70b, Gemini-3-f-p) for task 4. (LLM judge prompts are in Appendix \ref{app:prompt}).

    $\bullet$ \textbf{0.6-sim}: This judge method compared the semantic equivalence between the ground truth and the LLM answer in a cosine-similarity way: if the computed similarity value is above 0.6 (our configured empirical threshold), then the judge believes the two answers matched.
    
    $\bullet$ \textbf{Llama3.3-70b}: This judge (the LLM model), is prompted to decide whether the ground truth and the LLM answer are semantically equivalent. 
    
     $\bullet$ \textbf{Gemini-3-f-p}: This judge is used for task 4, since comparing two path(s) sets is more difficult, and we observed Gemini-3-f-p performs almost the best in all tasks (Table \ref{tab:task-difficulty}).

\subsection{Protocol Dataset}
\label{subsec:dataset}

We carefully selected 16 protocols covering Transport, Session, Application, and Data Link layers, including all the 14 protocols in \cite{PSMBench} and two more, spanning transport protocols (TCP \cite{rfctcp}, SCTP \cite{rfcsctp} and DCCP \cite{rfcdccp}), network configuration and control protocols (DHCP \cite{rfcdhcp}), tunneling and link protocols(PPP \cite{rfcppp} and PPTP \cite{rfcpptp}), application service protocols(FTP \cite{rfcftp} and NNTP \cite{rfcnntp}), congestion control protocols (TCP New Reno Congestion Control \cite{rfccgc2,rfccgc1}) email protocols (SMTP \cite{rfcsmtp}, IMAP \cite{rfcimap4} and POP3 \cite{rfcpop3}), real-time communication protocols (RTSP \cite{rfcrtsp} and SIP \cite{rfcsip}), routing protocols (BGP-4 \cite{rfcbgp}), and IoT-related protocols (MQTT \cite{rfcmqtt}). This results in a total sum of 118 states, 334 state transitions, 5381 valid paths, and therefore 118 queries in task 1, 235 queries in task 2, 255 queries in task 3, and 874 queries in task 4. Table \ref{tab:protocol-properties} displays the protocol collection we used in our evaluation, including the characteristics of each protocol. All gold FSMs, contexts, task queries, model outputs, judge outputs, and evaluation code are already publicly available.

\label{subsec:protocols}
\begin{table*}[htbp]
\centering
\small
\caption{Protocol characteristics and task workload summary. \#States and \#Trans describe the extracted FSM size. \#Q is the number of queries evaluated per protocol for each task. Ranked from low to high by the average score of tasks 1,2,3,4: a smaller number means harder.}
\label{tab:protocol-properties}
\begin{tabular}{p{1.1cm}p{0.7cm}p{0.7cm}p{0.8cm}p{1.5cm}p{1.5cm}p{2.3cm}p{0.5cm}p{0.5cm}p{0.5cm}p{1.2cm}}
\toprule
Protocol & \#States & \#Trans (Task1 \#Q) & \#Valid paths & Avg. reachable path length  & Avg. reachable paths set size   & FSM presentation in RFC & Task2 \#Q & Task3 \#Q & Task4 \#Q & Avg. Score (task1-4)\\
\midrule
RTSP       & 3 & 33 &36  &4.28   &6.00     & FSM table & 8 & 22        &6  &0.25\\
MQTT         & 12 & 17&115  &8.93  &1.53    & No state terms & 17 & 17   &132  &0.33 \\
PPP          & 10 &34 &3001  &12.49 &33.34  & Harder FSM table & 22 & 28 & 90 &0.34\\
TCP\_nr  & 4 & 15 &26   &5.00   &2.17   & Partial state terms & 12 & 11   &12  &0.37 \\
FTP          & 10 &24 &182  &6.66  &3.57   & No state terms & 19 & 17    &90 &0.37\\
BGP          & 6 & 24 &563  &8.16  &18.77  & Clear text FSM  & 15 & 17   &30  & 0.39\\
DCCP         & 9 & 21 &321  &9.43  &4.46  & ASCII FSM & 17 & 15          &72  &0.39 \\
NNTP         & 9& 14  &56  &6.43   &1.56   & No state terms & 12& 12     &72  &0.39 \\
PPTP         & 9& 21  &336  &7.95  &4.67   & ASCII FSM & 20 & 19         &72 &0.41 \\
DHCP         & 8 & 19 &84   &7.48  &1.95   & ASCII FSM  & 16 & 14        &56  & 0.43\\
SCTP         & 8 & 27 &426  &8.66 &7.61 & ASCII FSM & 19 & 14            &56  & 0.45\\
SMTP\_s  & 6 & 23 &57   &5.88   &2.71   & No state terms & 17 & 17       &30  &0.47 \\
SIP          & 5 & 20 &43  &5.09   &4.3   & ASCII FSM & 11 & 15          &20  & 0.51\\
POP3         & 3 & 18 &6   &3.67  &2.0  & Has state terms & 6 & 17       &6  & 0.54\\
TCP          & 11 &15 &107  &7.50  &2.38   & ASCII FSM & 15 & 13         &110  &0.57 \\
IMAP         & 5 & 9  &22  &4.73  &2.0   & ASCII FSM & 9 & 7             & 20 & 0.59\\

\bottomrule
\end{tabular}
\end{table*}

\section{Evaluation}
%
%
%
%
%
%
%
%
%
We evaluated 8 LLMs, including two open-source models (Deepseek-r1-distilled-qwen-32B, Llama3.3-70B) and 6 commercial LLMs (GPT-4o-mini, Deepseek-r1, Grok-3-mini, GPT-5.2, Gemini-3-flash-preview, Cluade-sonnet-4.5) to cover different vendors and parameter sizes. We deployed the open-source models with VLLM locally using NVIDIA H100 GPUs, with bfloat16 precision and a max context length of 8192. We used OpenRouter for all the commercial LLM API usage. We set the temperature to 0 for all LLM conversations to ensure the reasoning tasks' answers are deterministic and stable with the greedy decoding strategy. We also did bootstrap experiments to analyse the results' statistical certainty in Appendix \ref{app:stats}. We configured \texttt{length of generated response} to be capped at 1000 tokens as a trade-off of utility and token cost. We have a case study of how the output token limit affects the performance in Appendix \ref{app:outlimit} (Note that reasoning-token budgets are not exposed through the evaluated APIs and therefore cannot be controlled directly).

We explore the following questions:

 $\bullet$ \textbf{Q1} \textit{Whether different judges exhibit systematic scoring differences?}
 
 $\bullet$ \textbf{Q2} \textit{Which task is more difficult?}
 
 $\bullet$ \textbf{Q3} \textit{How do different context types affect each task's performance?}
 
 $\bullet$ \textbf{Q4} \textit{Which protocols are more challenging?}
 
 $\bullet$ \textbf{Q5} \textit{How do different LLMs perform?}

\textbf{Metrics.}
\label{subsec:sts}
Unless otherwise specified, all metrics defined below are computed
\emph{within a fixed task and a fixed answer-generating LLM}.
That is, for each \emph{task--LLM} pair, we analyze performance across protocols, context types, task metrics, and judges, while holding the task definition and the answer-generating LLM constant.

\paragraph{Notation.}
Let $\mathcal{P}$, $\mathcal{C}$, $\mathcal{M}$, and $\mathcal{J}$ denote the sets of protocols, context types (including \emph{Easy}, \emph{Manual-extracted-RFC}, \emph{LLM-extracted-RFC}, and \emph{No-context}), metrics (for task 1, $\mathcal{M}=\{s', A', \text{both}\}$, for task 2, $\mathcal{M}=\{C'\}$, for task 3, $\mathcal{M}=\{S_i\}$, and for task 4, $\mathcal{M}=\{P'\}$,  as shown in Table \ref{tab:task-difficulty}), and judges (0.6-sim, llama3.3-70b, gemini-3-f-p).

We treat each evaluation event $(p,c,m,j)$ as one observation and compute macro-averages by giving equal weight to each event. For each evaluation event of tuple $(p,c,m,j)\in \mathcal{P}\times\mathcal{C}\times\mathcal{M}\times\mathcal{J}$, we record the number of correct predictions $\mathrm{Correct}(p,c,m,j)$ and the number of queries $\mathrm{Queries}(p,c,m,j)$. We define the event-level accuracy as:
\begin{equation}
\mathrm{Acc}(p,c,m,j) \;=\; \frac{\mathrm{Correct}(p,c,m,j)}{\mathrm{Queries}(p,c,m,j)}.
\end{equation}

\paragraph{Q1 Judge score.}
To quantify the overall strictness of a judge, for each $j\in\mathcal{J}$ we compute:
\begin{equation}
\mathrm{Score}_{\text{j}}(j)
\;=\;
\frac{\sum_{p\in\mathcal{P}, c\in\mathcal{C}, m\in\mathcal{M}}
\mathrm{Acc}(p,c,m,j)}{|\mathcal{P}||\mathcal{C}||\mathcal{M}|}
.
\end{equation}

\paragraph{Q2 Task Difficulty Overall score.}
For the final overall score for a \textit{task-LLM} pair, we average by protocols, context types, metrics, and judges.
\begin{equation}
\mathrm{Score}_o =
\frac{
\sum_{p \in \mathcal{P},\, c \in \mathcal{C},\, m \in \mathcal{M},\, j \in \mathcal{J}}
\mathrm{Acc}(p,c,m,j)
}{
|\mathcal{P}|\,|\mathcal{C}|\,|\mathcal{M}|\,|\mathcal{J}|
}
\end{equation}


\paragraph{Q3 Context Effect.} For each context type $c\in\mathcal{C}$, we compute by averaging protocols, metrics and judges:
\begin{equation}
\mathrm{Score}_{\text{c}}(c)
\;=\;
\frac{\sum_{p\in\mathcal{P}, m\in\mathcal{M}, j\in\mathcal{J}}
\mathrm{Acc}(p,c,m,j)}{|\mathcal{P}||\mathcal{M}||\mathcal{J}|}.
\end{equation}

\paragraph{Q4 Protocol Difficulty.} For each protocol $p\in\mathcal{P}$, we average context types, metrics and judges:
\begin{equation}
\mathrm{Score}_{\text{p}}(p)
\;=\;
\frac{\sum_{c\in\mathcal{C}, m\in\mathcal{M}, j\in\mathcal{J}}\mathrm{Acc}(p,c,m,j)}{|\mathcal{C}||\mathcal{M}||\mathcal{J}|}
.
\end{equation}






\begin{table*}[ht]
\centering
\small
\caption{Judge score on Task~3 across answering LLMs.}
\label{tab:task3-overall}
\begin{tabular}{p{2cm}p{1cm}p{1cm}p{1cm}p{1cm}p{1cm}p{1cm}p{1cm}p{1cm}}
\toprule
Judge & gpt-4o-mini & qwen32B  & llama3.3-70B  & deepseek-r1 & grok-3-mini & gpt-5.2 & gemini-3-f-p & claude-s-4.5  \\
\midrule
Sim-0.6      & 0.23 & 0.37 & 0.43   &0.53  &0.56&0.63  & 0.60 &0.58\\
Llama3.3-70b & 0.17 & 0.28  &0.17   &0.44  &0.46 &0.52 & 0.49 &0.49 \\

\bottomrule
\end{tabular}
\end{table*}

\subsection{Q1 Results: Judge Bias}
Past work~\cite{PSMBench} used a cosine-similarity-based judge to compare the LLM-generated answers with the ground truth. We argue that since this is essentially a paraphrase detection task, relying on an LLM would be more reliable (as demonstrated in Appendix \ref{app:llm-human}) and analyze the differences between the two judge types. 

For both Task 1 and 2, we observed the same trend among judges (Tab. \ref{tab:task1-overall}, \ref{tab:task2-overall}): the 0.6-sim judge produces a lower score, and the Llama3.3-70B judge is slightly more lenient. This can be because LLM judges can detect different phrases conveying the same semantics, while the cosine similarity judges cannot, e.g., the LLM judge will decide \texttt{Ground Truth:} \texttt{Request token from AS} and \texttt{LLM:} \texttt{Client sends token request to Authorization Server (AS) via token endpoint  (Figure 1’, ’step A) } equivalent, while the 0.6-sim judge will not.

On Task 3, the opposite trend was observed: the sim-0.6 judge is more lenient and consistently produces higher scores than the LLM judge (Tab. \ref{tab:task3-overall}). We observed that when evaluating \textit{previous state(s)} answers, which are usually short terms, the cosine-similarity-based judge may treat two short state phrases with similar wording as equivalent, even though they actually refer to different protocol states. For example, the 0.6-sim judge will decide \texttt{Ground Truth Previous State: Connected} and \texttt{LLM Previous State: Disconnected} equivalent, while the LLM judge will not.
For Task 4, we experimented with several cosine-similarity thresholds, 0.6, 0.7, and 0.8 (see Appendix \ref{app:t4gemini}), but all were still inconsistent with the LLM judges. So, for Task 4, we only relied on the two LLM judges, Gemini-3-flash-preview and Llama3.3-70b.


\begin{table*}[ht]
\centering
\small
\caption{Task difficulty on Task~1, 2, 3, 4 (smaller number is harder).}
\label{tab:task-difficulty}
\begin{tabular}{p{3.5cm}p{1cm}p{1cm}p{1cm}p{1cm}p{1cm}p{0.5cm}p{1cm}p{1cm}p{0.5cm}}
\toprule
Metric 
& gpt-4o-mini & qwen32B & llama3.3-70B
 & deepseek-r1& grok-3-mini & gpt-5.2 & gemini-3-f-p & claude-s-4.5 & \textbf{Avg.} \\
\midrule
$s'$ (Task1 next state) &0.74 & 0.71  & 0.78  & 0.81 & 0.82&0.84 & 0.85 &  0.87  & 0.80\\
$A'$ (Task1 actions) &0.38 & 0.41  & 0.44  & 0.48 & 0.48&0.49  & 0.50 & 0.51 & 0.46\\
Both (Task1 both) &0.34 & 0.37  & 0.41  & 0.45 & 0.44&0.46  & 0.48 & 0.50 &0.43 \\
$C'$ (Task2 condition(s)) &0.30 & 0.33  & 0.32  & 0.34 &0.40 &0.35 & 0.42 &  0.42 & 0.36\\
$S_i$(Task3 previous state(s)) &0.20  & 0.32 & 0.38  & 0.48 &0.51 &0.58 & 0.54 &  0.54 &0.44\\
$P'$ (Task4 path(s) set) & 0.06 &0.33 &0.27 &0.39 &0.38 &0.41 &0.34 &0.37 &0.32 \\
\textbf{Avg.} &0.34 &0.41 &0.43 &0.49 &0.51  &0.52 &0.52 &0.54 & 0.47 \\
\bottomrule
\end{tabular}
\end{table*}

\subsection{Q2 Results: Task Difficulty}
\label{subsec:taskdiff}
In Table \ref{tab:task-difficulty}, a lower score indicates a greater task difficulty. In Task 1, simply predicting \textit{Next State} is the easiest (avg. accuracy 0.80), and predicting \textit{Actions} is harder (avg. accuracy 0.46), while predicting \textit{both} is the hardest (avg. accuracy 0.43).

The average accuracy (0.36) on Task~2 (\textit{Condition} Inference) is even lower than Task~1(\textit{both}), indicating that Task~2 is inherently more challenging. The average accuracy (0.44) on Task~3 is comparable to Task~1(\textit{both}); but compared with Task~2, some weak LLMs (gpt-4o-mini, qwen32B) found Task~3 harder, while other stronger LLMs found the opposite. This revealed that capable LLMs kind of overcame the difficulty of retrieving initial states. Task~4 produced the lowest accuracy of 0.32.


\paragraph{Difficulty Analysis.}
The four tasks evaluate distinct aspects of FSM understanding and vary substantially in reasoning complexity.

Task 1 (Next-state-action Prediction) primarily requires identifying the correct outgoing transition from the current state. Performance is affected by the number of candidate transitions available from a state, ambiguity caused by overlapping transition conditions, and uneven representation of individual transition rules.

Task 2 (Condition Inference) requires recovering the triggering condition of a state transition. Compared to Task 1, the target output is a natural-language condition rather than a discrete state label, introducing additional challenges in sentence-level semantic understanding, condition boundary identification, and distinguishing among multiple conditions that lead to the same transition.

Task 3 (Previous-state Inference) requires reasoning in the reverse direction of the FSM. Models must infer which preceding state and transition condition could have produced the observed outcome. This task combines condition matching with backward reasoning and becomes particularly difficult when multiple predecessor states share similar transition outcomes.

Task 4 (Path Prediction) requires constructing and traversing an implicit graph representation of the FSM. Difficulty increases with path length and the number of alternative paths connecting two states. Successful performance depends not only on understanding individual transitions but also on maintaining a coherent graph structure and performing multi-step reasoning over it.

\paragraph{Reasoning Complexity Across Tasks.}
Task 1 and Task 2 achieve near-ceiling performance (Table \ref{tab:task1-context-effect}, \ref{tab:task2-context-effect}) under structured FSM representations (Easy-context), suggesting that modern LLMs can reliably perform transition selection and condition identification when relevant transition rules are explicitly provided. Performance drops for Task 3 (Table \ref{tab:task3-context-effect}), which requires reverse reasoning over state transitions rather than forward prediction. Task 4 remains the most difficult task despite the availability of structured FSM representations (Table \ref{tab:task4-context-effect}), indicating that multi-hop path reasoning introduces additional challenges beyond local transition understanding. Solving Task 4 requires constructing an internal graph representation, maintaining connectivity information across multiple states, and exploring alternative paths, substantially increasing the reasoning search space.

\subsection{Q3 Results: Context Effect}
We evaluated how different contexts would affect the task performance, and we have the results table for each task (Table \ref{tab:task1-context-effect}, \ref{tab:task2-context-effect}, \ref{tab:task3-context-effect}, \ref{tab:task4-context-effect}). As expected, the \emph{easy} context, i.e., a human-organized transition table, produced the highest score, almost above 90\% for tasks 1 and 2; but for tasks 3 and 4, due to the inherent difficulty of tasks, even with the \emph{easy} context, LLM's score dropped below 90\% and 80\%. Across all tasks, the \emph{Manual-extracted-RFC} context mostly outperformed the \emph{LLM-extracted-RFC} context (only a few were equal), showing that manually reviewing documents to select relevant material benefits task performance.

The gap between \emph{LLM-extracted-RFC} and \emph{No-context} is close: mostly, the former produced better results, but we noticed that for tasks 2 and 4 (Table \ref{tab:task2-context-effect}, \ref{tab:task4-context-effect}), a few cases show \emph{No-context} outperformed.

\noindent\textbf{Challenges in Condition Extraction.}
In Table \ref{tab:task1-context-effect}, \ref{tab:task2-context-effect}, \ref{tab:task3-context-effect}, \ref{tab:task4-context-effect}, focusing on the rows for Manual-extracted-RFC and LLM-extracted-RFC, we found that task 2's performance drops the most sharply from Easy-context, compared with other tasks: Task 2 requires identifying and reproducing transition conditions, which are often expressed as abstract logical statements embedded within a long RFC description. Unlike Tasks 1 and 3, where the output space consists of discrete state labels, Task 2 requires locating precise condition boundaries while excluding surrounding irrelevant text. As a result, errors frequently arise from including irrelevant contextual information or extracting incomplete conditions (e.g., missing B from ``A and B"). Task 4 shows the same trend because it is also affected by the same factor: the path(s) answer should include both the condition and the state, but correctly extracting both from the long documents is difficult.

\begin{table*}[ht]
\centering
\small
\caption{Effect of context type on Task~2 (smaller number is harder).}
\label{tab:task2-context-effect}
\begin{tabular}
{p{3cm}p{1cm}p{1cm}p{1cm}p{1cm}p{1cm}p{1cm}p{1cm}p{1cm}}
\toprule
Context Type & gpt-4o-mini & qwen32B  & llama3.3-70B & deepseek-r1& grok-3-mini & gpt-5.2 & gemini-3-f-p & claude-s-4.5  \\
\midrule
Easy                 & 0.69  & 0.90  & 0.87   & 0.76  &0.92 &0.89  & 0.92 & 0.92 \\
Manual-extracted-RFC & 0.23  & 0.19  & 0.19   & 0.25  &0.26 &0.23  &0.29 &0.28   \\
LLM-extracted-RFC    & 0.19  & 0.12  &  0.15  & 0.17  &0.20 &0.15 & 0.25 &0.22   \\
No-context           &0.11   & 0.09  & 0.07   &0.20   &0.21 &0.14 & 0.22 &0.24    \\
\bottomrule
\end{tabular}
\end{table*}

\subsection{Q4 Results: Protocol Difficulty}
We rank LLMs based on their overall performance at the bottom row in Tab.~\ref{tab:task-difficulty}, and performance varies for each LLM by protocol. Given our comprehensive collection of protocols (Section~\ref {subsec:dataset}), we explore the factors that could explain performance variation across the different protocols.

\begin{figure}[htbp]
    \centering    \includegraphics[width=\linewidth]{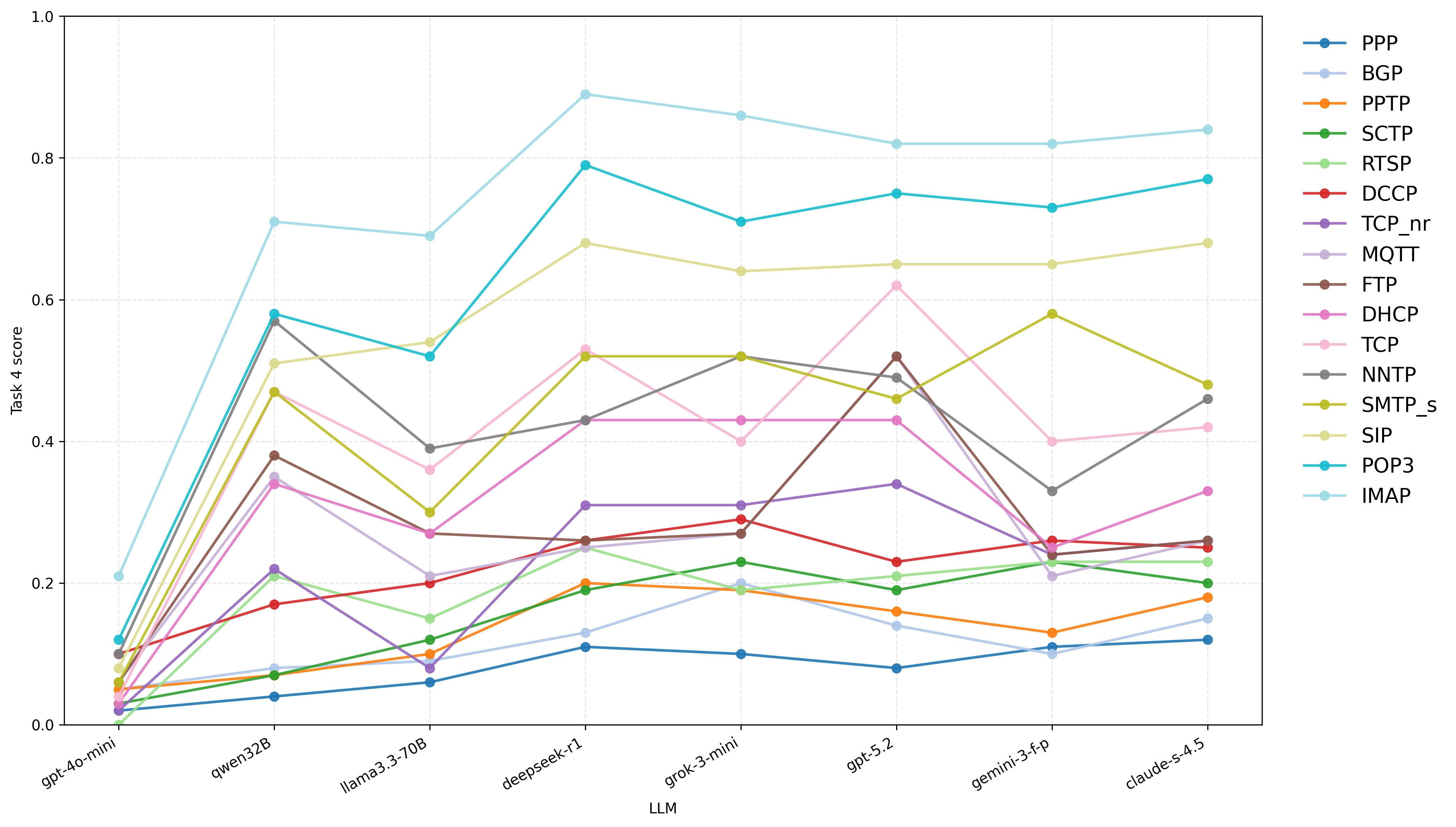}
    \caption{Per-protocol-llm Performance on Task 4.}
    \label{fig:p4}
\end{figure}

The analysis, summarized in Tables~\ref{tab:task1-protocol-difficulty}, \ref{tab:task2-protocol-difficulty}, \ref{tab:task3-protocol-difficulty}, \ref{tab:task4-protocol-difficulty}, and corresponding figures (Fig.~\ref{fig:p1}, \ref{fig:p2}, \ref{fig:p3}, \ref{fig:p4}), shows that protocol difficulty rankings are different from task to task, suggesting that protocol-specific characteristics and task-specific factors both affect the results. 
For example, in Table ~\ref{tab:protocol-properties}, we notice that the protocols (IMAP, TCP, SIP) having the highest scores (averaged by all tasks) tend to have an ASCII FSM in the RFC (example in Appendix \ref{app:rfc}), making the text comprehension task significantly easier as all relevant information is concisely summarized in a focused way, compared to RFCs that require understanding verbose text, retrieved from multiple locations, or understanding table-formatted FSMs (example in Appendix \ref{app:rfc}), such as in RTSP, MQTT, PPP.


RTSP is an interesting case: for tasks 1 and 2, it got the worst score, while for tasks 3 and 4, it's ranked in the middle. From Table \ref{tab:protocol-properties}, we know that it has only 3 states (least among all protocols), but 33 transitions (2nd highest), creating an asymmetric difficulty between tasks 1 and 2 (reasoning over 33 transitions, expressed in a challenging table format), and task 3, which can be trivially solved.

 Task 4 per-protocol-llm performance is also interesting. In Fig. \ref{fig:p4}, there are clear gaps between protocol lines, showing the inherent difficulty of protocols under the same task. When we refer to the characteristics of the protocols (Table \ref{tab:protocol-properties}), the observed gaps can be understood: protocols that have the least valid paths, average reachable path length and average reachable path set size have the best performance (IMAP, POP3, SIP), while the protocols that have a large number of valid paths, average reachable path length and average reachable path set size produced the worst performance (PPP, BGP, PPTP). When exploration breadth and depth are both extended, LLMs will find the task harder to solve. Also, from Table \ref{tab:task4-protocol-difficulty}, we found IMAP, POP3, and SIP exhibit the highest average scores but also the largest standard deviations, indicating substantial performance gaps between weaker and stronger models even in those easy cases.

\subsection{Q5 Results: Different LLMs' Performance }
In Table \ref{tab:task-difficulty}, the last row shows the LLM performance score averaged across tasks, and we put from left to right an increasing LLM performance. For a fine-grained per-task-protocol view, we can also track the plot lines in Fig. \ref{fig:p1}, \ref{fig:p2}, \ref{fig:p3}, \ref{fig:p4}: in most of them, a trend of increasing LLM performance can be observed from left to right: with GPT-4o-mini (estimated to have the minial parameter size) the worst, and Claude-sonnet-4.5 the best. 

Interestingly, DeepSeek-r1 and Grok-3-mini outperformed Gemini-3-f-p and Claude-s-4.5 on Task 4 (Tab. \ref{tab:task-difficulty}), showing that the LLMs' claimed deep reasoning abilities may contribute to task~4 that requires deep search both in depth and breadth.

\subsection{Task4 Precision and Recall Analysis}
As shown in subsection \ref{subsec:taskdiff}, task 4 is the hardest based on a perfect path-set match accuracy. To understand the partial-match performance, we compute the precision, recall, and F1 (Tab. \ref{tab:task4-protocol-prf}). 

Consider one LLM answering the Task 4 queries with a context-type $c$ and a fixed judge $j$, for a given protocol $p$, let $X_p$ denote the number of queries.
For each query $i$, we define:
%

$\bullet$ \textbf{$GT_i$}: the set of ground-truth paths for query $i$, which keeps the same across contexts.

    $\bullet$ \textbf{$Ans_{i,c}$}: the set of paths predicted by the LLM for query $i$ and context $c$.
    
    $\bullet$ \textbf{$recall\_hits_{i,c}$}: the ground-truth paths covered by the LLM prediction for query $i$ and context $c$.
    
    $\bullet$ \textbf{$precision\_hits_{i,c}$}: the number of valid LLM-predicted paths (i.e., that match ground truth) for query $i$ and context $c$.

We first accumulate statistics across all queries within protocol p:
$total\_recall\_hits_{p,c} = \sum_{i=1}^{X_p}
recall\_hits_{i,c}$, $total\_precision\_hits_{p,c}
= \sum_{i=1}^{X_p} precision\_hits_{i,c}$, $total\_GT_p
= \sum_{i=1}^{X_p}
|GT_i|$, $total\_Pred_{p,c}
= \sum_{i=1}^{X_p}|Ans_{ic}|$.

\noindent With that, we compute the Micro R/P/F1. 

We have the $Recall_{p,c}$, $Precision_{p,c}$, and $F1_{p,c}$ for each context type (full details of the formulas to compute R/P/F1 are in Appendix \ref{app:formulas}.), and then we computed $Recall_p$, $Precision_p$, and $F1_p$ for each protocol by averaging between the context types. We averaged between the two judges (Gemini-3-flash-preview and Llama3.3-70b) to get the average R/P/F1 for each protocol. From Table \ref{tab:task4-protocol-prf}, we observe that for most protocols, the precision is higher than the recall, which reveals that most errors come from the LLM not outputting enough valid answers, instead of the LLM over-generating invalid paths.

\section{Conclusion}

 This paper assesses the extent to which LLMs can interpret protocol specifications correctly from textual descriptions. We designed 4 tasks and 1,482 task queries for 16 protocols. We evaluated different judge biases, the inherent difficulty gaps between tasks, different context types, and the influence of protocol characteristics. Our work represents a step toward verifying whether LLMs can be trusted in FSM reasoning out of protocol specifications.

\section*{Limitations}
Our work focused on designing the 4 tasks to evaluate LLMs' understanding of the FSM logic expressed in the protocol RFC documents. FSM is an important factor that guides the execution of a protocol, but other factors related to the protocol's design also matter, such as message packet structure details, error-handling behavior, valid configuration, etc., which, for now, are out of our evaluation. Besides, we didn't consider any implementation choices or real-world code base mapping of the textual definition of protocol design, which can be a future work. Our conclusions are limited to mature IETF protocols researched in this work.

\section*{Ethical considerations}
The goal of our work is
to evaluate different LLMs' reasoning abilities in the FSM logic of different network protocols, providing different context types like processed-RFC documents, etc. We collected our data from publicly available RFCs. Our prompt questions are technical reasoning tasks and serve no harm to other parties. Benchmark scores should not be interpreted as vendor rankings or as evidence that current LLMs are ready for security-critical protocol analysis without human oversight. We have open-sourced our code at \url{https://anonymous.4open.science/r/rfcllm}.

\section*{Acknowledgments}

\bibliography{rfcllm}

@inproceedings{chatafl,
author={Ruijie Meng and Martin Mirchev and Marcel B\"{o}hme and Abhik Roychoudhury},
title={Large Language Model guided Protocol Fuzzing},
booktitle={Proceedings of the 31st Annual Network and Distributed System Security Symposium (NDSS)},
year={2024},
url={https://www.ndss-symposium.org/ndss-paper/large-language-model-guided-protocol-fuzzing/}

}

@inproceedings{llmdreamsockets,
author = {Arkko, Jari and Lindbo, Dag and Klitte, Martin},
title = {Do Large Language Models Dream of Sockets?},
year = {2024},
isbn = {9798400707230},
publisher = {Association for Computing Machinery},
address = {New York, NY, USA},
url = {https://doi.org/10.1145/3673422.3674900},
doi = {10.1145/3673422.3674900},
booktitle = {Proceedings of the 2024 Applied Networking Research Workshop},
pages = {103–105},
numpages = {3},
location = {Vancouver, AA, Canada},
series = {ANRW '24}
}

@inproceedings{tcpwn_ndss2018,
author = {Jero, Samuel and Hoque, Endadul and Choffnes, David and Mislove, Alan and Nita-Rotaru, Cristina},
title = {Automated Attack Discovery in TCP Congestion Control Using a Model-guided Approach},
year = {2018},
isbn = {9781450355858},
publisher = {Association for Computing Machinery},
address = {New York, NY, USA},
url = {https://doi.org/10.1145/3232755.3232769},
doi = {10.1145/3232755.3232769},
booktitle = {Proceedings of the 2018 Applied Networking Research Workshop},
pages = {95},
numpages = {1},
location = {Montreal, QC, Canada},
series = {ANRW '18}
}

@misc{rfccgc1,
    series =    {Request for Comment},
    number =    5681,
    howpublished =  {RFC 5681},
    publisher = {RFC Editor},
    doi =       {10.17487/RFC5681},
    url =       {https://www.rfc-editor.org/info/rfc5681},
    author =    {Ethan Blanton and Dr. Vern Paxson and Mark Allman},
    title =     {{TCP Congestion Control}},
    pagetotal = 18,
    year =      2009,
    month =     sep,

}

@misc{rfccgc2,
    series =    {Request for Comments},
    number =    6582,
    howpublished =  {RFC 6582},
    publisher = {RFC Editor},
    doi =       {10.17487/RFC6582},
    url =       {https://www.rfc-editor.org/info/rfc6582},
    author =    {Andrei Gurtov and Tom Henderson and Sally Floyd and Yoshifumi Nishida},
    title =     {{The NewReno Modification to TCP's Fast Recovery Algorithm}},
    pagetotal = 16,
    year =      2012,
    month =     apr,

}

@misc{rfcppp,
    series =    {Request for Comments},
    number =    1661,
    howpublished =  {RFC 1661},
    publisher = {RFC Editor},
    doi =       {10.17487/RFC1661},
    url =       {https://www.rfc-editor.org/info/rfc1661},
    author =    {William A. Simpson},
    title =     {{The Point-to-Point Protocol (PPP)}},
    pagetotal = 53,
    year =      1994,
    month =     jul,
  
}

@misc{rfcmqtt,
    series =    {Request for Comments},
    number =    9431,
    howpublished =  {RFC 9431},
    publisher = {RFC Editor},
    doi =       {10.17487/RFC9431},
    url =       {https://www.rfc-editor.org/info/rfc9431},
    author =    {Cigdem Sengul and Anthony Kirby},
    title =     {{Message Queuing Telemetry Transport (MQTT) and Transport Layer Security (TLS) Profile of Authentication and Authorization for Constrained Environments (ACE) Framework}},
    pagetotal = 33,
    year =      2023,
    month =     jul,
}

@misc{rfcsmtp,
    series =    {Request for Comments},
    number =    5321,
    howpublished =  {RFC 5321},
    publisher = {RFC Editor},
    doi =       {10.17487/RFC5321},
    url =       {https://www.rfc-editor.org/info/rfc5321},
    author =    {Dr. John C. Klensin},
    title =     {{Simple Mail Transfer Protocol}},
    pagetotal = 95,
    year =      2008,
    month =     oct
}

@misc{rfcsip,
    series =    {Request for Comments},
    number =    3261,
    howpublished =  {RFC 3261},
    publisher = {RFC Editor},
    doi =       {10.17487/RFC3261},
    url =       {https://www.rfc-editor.org/info/rfc3261},
    author =    {Eve Schooler and Jonathan Rosenberg and Henning Schulzrinne and Alan Johnston and Gonzalo Camarillo and Jon Peterson and Robert Sparks and Mark J. Handley},
    title =     {{SIP: Session Initiation Protocol}},
    pagetotal = 269,
    year =      2002,
    month =     jul
}

@misc{rfcrtsp,
    series =    {Request for Comments},
    number =    7826,
    howpublished =  {RFC 7826},
    publisher = {RFC Editor},
    doi =       {10.17487/RFC7826},
    url =       {https://www.rfc-editor.org/info/rfc7826/},
    author =    {H. Schulzrinne and A. Rao and R. Lanphier and M. Westerlund and M. Stiemerling},
    title =     {{Real-Time Streaming Protocol Version 2.0}},
    pagetotal = 318,
    year =      2016,
    month =     dec,
}

@misc{rfcdccp,
    series =    {Request for Comments},
    number =    4340,
    howpublished =  {RFC 4340},
    publisher = {RFC Editor},
    doi =       {10.17487/RFC4340},
    url =       {https://www.rfc-editor.org/info/rfc4340},
    author =    {Sally Floyd and Mark J. Handley and Eddie Kohler},
    title =     {{Datagram Congestion Control Protocol (DCCP)}},
    pagetotal = 129,
    year =      2006,
    month =     mar,
}

@misc{rfcsctp,
    series =    {Request for Comments},
    number =    4960,
    howpublished =  {RFC 4960},
    publisher = {RFC Editor},
    doi =       {10.17487/RFC4960},
    url =       {https://www.rfc-editor.org/info/rfc4960},
    author =    {Randall R. Stewart},
    title =     {{Stream Control Transmission Protocol}},
    pagetotal = 152,
    year =      2007,
    month =     sep,

}

@misc{rfcbgp,
    series =    {Request for Comments},
    number =    4271,
    howpublished =  {RFC 4271},
    publisher = {RFC Editor},
    doi =       {10.17487/RFC4271},
    url =       {https://www.rfc-editor.org/info/rfc4271},
    author =    {Yakov Rekhter and Susan Hares and Tony Li},
    title =     {{A Border Gateway Protocol 4 (BGP-4)}},
    pagetotal = 104,
    year =      2006,
    month =     jan,
}

@misc{rfcdhcp,
    series =    {Request for Comments},
    number =    2131,
    howpublished =  {RFC 2131},
    publisher = {RFC Editor},
    doi =       {10.17487/RFC2131},
    url =       {https://www.rfc-editor.org/info/rfc2131},
    author =    {Ralph Droms},
    title =     {{Dynamic Host Configuration Protocol}},
    pagetotal = 45,
    year =      1997,
    month =     mar,
}

@misc{rfctcp,
    series =    {Request for Comments},
    number =    9293,
    howpublished =  {RFC 9293},
    publisher = {RFC Editor},
    doi =       {10.17487/RFC9293},
    url =       {https://www.rfc-editor.org/info/rfc9293},
    author =    {Wesley Eddy},
    title =     {{Transmission Control Protocol (TCP)}},
    pagetotal = 98,
    year =      2022,
    month =     aug,
}

@INPROCEEDINGS{rfcaudit,
  author={Zheng, Mingwei and Wang, Chengpeng and Liu, Xuwei and Guo, Jinyao and Feng, Shiwei and Zhang, Xiangyu},
  booktitle={2025 40th IEEE/ACM International Conference on Automated Software Engineering (ASE)}, 
  title={RFCAudit: AI Agent for Auditing Protocol Implementations Against RFC Specifications}, 
  year={2025},
  volume={},
  number={},
  pages={1221-1233},
  doi={10.1109/ASE63991.2025.00105}}

@misc{llmtemporal,
      title={nl2spec: Interactively Translating Unstructured Natural Language to Temporal Logics with Large Language Models}, 
      author={Matthias Cosler and Christopher Hahn and Daniel Mendoza and Frederik Schmitt and Caroline Trippel},
      year={2023},
      eprint={2303.04864},
      archivePrefix={arXiv},
      primaryClass={cs.LO},
      url={https://arxiv.org/abs/2303.04864}, 
}

@misc{formalllm,
      title={Formal Specifications from Natural Language}, 
      author={Christopher Hahn and Frederik Schmitt and Julia J. Tillman and Niklas Metzger and Julian Siber and Bernd Finkbeiner},
      year={2022},
      eprint={2206.01962},
      archivePrefix={arXiv},
      primaryClass={cs.SE},
      url={https://arxiv.org/abs/2206.01962}, 
}

@INPROCEEDINGS{llmmodel,
  author={Mao, Ziyu and Wang, Jingyi and Sun, Jun and Qin, Shengchao and Xiong, Jiawen},
  booktitle={2025 IEEE/ACM 47th International Conference on Software Engineering (ICSE)}, 
  title={LLM-Aided Automatic Modeling for Security Protocol Verification}, 
  year={2025},
  volume={},
  number={},
  pages={642-654},
  doi={10.1109/ICSE55347.2025.00197}}

@misc{llmassistedmodelbasedfuzzingprotocol,
      title={LLM-Assisted Model-Based Fuzzing of Protocol Implementations}, 
      author={Changze Huang and Di Wang and Zhi Quan Zhou},
      year={2025},
      eprint={2508.01750},
      archivePrefix={arXiv},
      primaryClass={cs.CR},
      url={https://arxiv.org/abs/2508.01750}, 
}

@Article{statepre,
AUTHOR = {Zhang, Yifan and Zhu, Kailong and Peng, Jie and Lu, Yuliang and Chen, Qian and Li, Zixiong},
TITLE = {StatePre: A Large Language Model-Based State-Handling Method for Network Protocol Fuzzing},
JOURNAL = {Electronics},
VOLUME = {14},
YEAR = {2025},
NUMBER = {10},
ARTICLE-NUMBER = {1931},
URL = {https://www.mdpi.com/2079-9292/14/10/1931},
ISSN = {2079-9292},
DOI = {10.3390/electronics14101931}
}

@article{rfcnlp_sp2022,
  title={Automated Attack Synthesis by Extracting Finite State Machines from Protocol Specification Documents},
  author={Maria Leonor Pacheco and Max von Hippel and Ben Weintraub and Dan Goldwasser and Cristina Nita-Rotaru},
  journal={2022 IEEE Symposium on Security and Privacy (SP)},
  year={2022},
  pages={51-68},
  url={https://api.semanticscholar.org/CorpusID:247012059}
}

@inproceedings {llmiot,
author = {Xiaoyue Ma and Lannan Luo and Qiang Zeng},
title = {From One Thousand Pages of Specification to Unveiling Hidden Bugs: Large Language Model Assisted Fuzzing of Matter {IoT} Devices},
booktitle = {33rd USENIX Security Symposium (USENIX Security 24)},
year = {2024},
isbn = {978-1-939133-44-1},
address = {Philadelphia, PA},
pages = {4783--4800},
url = {https://www.usenix.org/conference/usenixsecurity24/presentation/ma-xiaoyue},
publisher = {USENIX Association},
month = aug
}

@inproceedings{sctp_2024,
  title     = {A Formal Analysis of SCTP: Attack Synthesis and Patch Verification},
  author    = {Ginesin, Jacob and von Hippel, Max and Defloor, Evan and Nita-Rotaru, Cristina and Tuxen, Michael},
  booktitle = {Proceedings of the 33rd USENIX Security Symposium (USENIX Security '24)},
  year      = {2024},
  publisher = {USENIX Association},
  url ={https://www.usenix.org/conference/usenixsecurity24/presentation/ginesin}
}

@inproceedings{prosper_hotnet2023,
author = {Sharma, Prakhar and Yegneswaran, Vinod},
title = {PROSPER: Extracting Protocol Specifications Using Large Language Models},
year = {2023},
isbn = {9798400704154},
publisher = {Association for Computing Machinery},
address = {New York, NY, USA},
url = {https://doi.org/10.1145/3626111.3628205},
doi = {10.1145/3626111.3628205},
booktitle = {Proceedings of the 22nd ACM Workshop on Hot Topics in Networks},
pages = {41–47},
numpages = {7},
location = {Cambridge, MA, USA},
series = {HotNets '23}
}

@misc{rfcftp,
    series =    {Request for Comments},
    number =    959,
    howpublished =  {RFC 959},
    publisher = {RFC Editor},
    doi =       {10.17487/RFC0959},
    url =       {https://www.rfc-editor.org/info/rfc959},
    author =    {},
    title =     {{File Transfer Protocol}},
    pagetotal = 69,
    year =      1985,
    month =     oct,

}

@misc{rfcpptp,
    series =    {Request for Comments},
    number =    2637,
    howpublished =  {RFC 2637},
    publisher = {RFC Editor},
    doi =       {10.17487/RFC2637},
    url =       {https://www.rfc-editor.org/info/rfc2637},
    author =    {Glen Zorn and Gurdeep-Singh Pall and Kory Hamzeh},
    title =     {{Point-to-Point Tunneling Protocol (PPTP)}},
    pagetotal = 57,
    year =      1999,
    month =     jul,
}

@misc{rfcimap4,
    series =    {Request for Comments},
    number =    9051,
    howpublished =  {RFC 9051},
    publisher = {RFC Editor},
    doi =       {10.17487/RFC9051},
    url =       {https://www.rfc-editor.org/info/rfc9051},
    author =    {Alexey Melnikov and Barry Leiba},
    title =     {{Internet Message Access Protocol (IMAP) - Version 4rev2}},
    pagetotal = 163,
    year =      2021,
    month =     aug,
}

@misc{rfcpop3,
    series =    {Request for Comments},
    number =    1939,
    howpublished =  {RFC 1939},
    publisher = {RFC Editor},
    doi =       {10.17487/RFC1939},
    url =       {https://www.rfc-editor.org/info/rfc1939},
    author =    {Dr. Marshall T. Rose and John G. Myers},
    title =     {{Post Office Protocol - Version 3}},
    pagetotal = 23,
    year =      1996,
    month =     may,
}

@misc{rfcnntp,
    series =    {Request for Comments},
    number =    3977,
    howpublished =  {RFC 3977},
    publisher = {RFC Editor},
    doi =       {10.17487/RFC3977},
    url =       {https://www.rfc-editor.org/info/rfc3977},
    author =    {Clive Feather},
    title =     {{Network News Transfer Protocol (NNTP)}},
    pagetotal = 125,
    year =      2006,
    month =     oct,
}

@inproceedings{PSMBench,
  title     = {PSMBench: A Benchmark and Dataset for Evaluating LLMs Extraction of Protocol State Machines from RFC Specifications},
  author    = {Shen, Zilin and Luo, Xinyu and Karim, Imtiaz and Bertino, Elisa},
  booktitle = {Advances in Neural Information Processing Systems (NeurIPS) Datasets and Benchmarks Track},
  year      = {2025},
  note      = {Poster},
  url       = {https://neurips.cc/virtual/2025/poster/121835}
}

@article{vafa2024evaluating,
  title={Evaluating the world model implicit in a generative model},
  author={Vafa, Keyon and Chen, Justin Y and Rambachan, Ashesh and Kleinberg, Jon and Mullainathan, Sendhil},
  journal={Advances in Neural Information Processing Systems},
  volume={37},
  pages={26941--26975},
  year={2024},
  url={https://dl.acm.org/doi/10.5555/3737916.3738762}
}

@inproceedings{zhang-etal-2024-llm-graph,
    title = "Can {LLM} Graph Reasoning Generalize beyond Pattern Memorization?",
    author = "Zhang, Yizhuo  and
      Wang, Heng  and
      Feng, Shangbin  and
      Tan, Zhaoxuan  and
      Han, Xiaochuang  and
      He, Tianxing  and
      Tsvetkov, Yulia",
    editor = "Al-Onaizan, Yaser  and
      Bansal, Mohit  and
      Chen, Yun-Nung",
    booktitle = "Findings of the Association for Computational Linguistics: EMNLP 2024",
    month = nov,
    year = "2024",
    address = "Miami, Florida, USA",
    publisher = "Association for Computational Linguistics",
    url = "https://aclanthology.org/2024.findings-emnlp.127/",
    doi = "10.18653/v1/2024.findings-emnlp.127",
    pages = "2289--2305"
}

@article{wang2023can,
  title={Can language models solve graph problems in natural language?},
  author={Wang, Heng and Feng, Shangbin and He, Tianxing and Tan, Zhaoxuan and Han, Xiaochuang and Tsvetkov, Yulia},
  journal={Advances in Neural Information Processing Systems},
  volume={36},
  pages={30840--30861},
  year={2023},
  url={https://dl.acm.org/doi/10.5555/3666122.3667467}
}

@inproceedings{yuan-etal-2025-gracore,
    title = "{G}ra{C}o{R}e: Benchmarking Graph Comprehension and Complex Reasoning in Large Language Models",
    author = "Yuan, Zike  and
      Liu, Ming  and
      Wang, Hui  and
      Qin, Bing",
    editor = "Rambow, Owen  and
      Wanner, Leo  and
      Apidianaki, Marianna  and
      Al-Khalifa, Hend  and
      Eugenio, Barbara Di  and
      Schockaert, Steven",
    booktitle = "Proceedings of the 31st International Conference on Computational Linguistics",
    month = jan,
    year = "2025",
    address = "Abu Dhabi, UAE",
    publisher = "Association for Computational Linguistics",
    url = "https://aclanthology.org/2025.coling-main.531/",
    pages = "7925--7948"
}

@inproceedings{snake_dsn2015,
  title     = {Leveraging State Information for Automated Attack Discovery in Transport Protocol Implementations},
  author    = {Jero, Samuel and Lee, Hyunwoo and Nita-Rotaru, Cristina},
  booktitle = {Proceedings of the 45th IEEE/IFIP International Conference on Dependable Systems and Networks (DSN)},
  year      = {2015},
  month     = jun,
  publisher = {IEEE},
  url ={https://dl.acm.org/doi/10.1109/DSN.2015.22}
}

\appendix

\section{RFC example}
\label{app:rfc}

\begin{figure}[t]
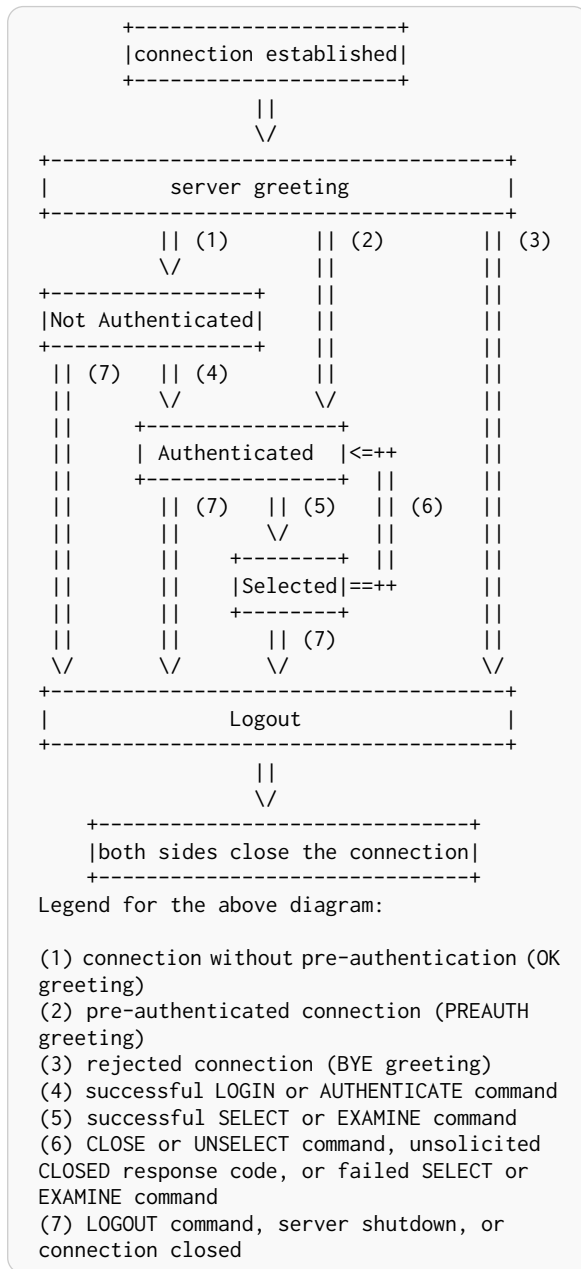

\centering
\begin{promptbox}
\begin{verbatim}
       +----------------------+
       |connection established|
       +----------------------+
                  ||
                  \/
+--------------------------------------+
|          server greeting             |
+--------------------------------------+
          || (1)       || (2)        || (3)
          \/           ||            ||
+-----------------+    ||            ||
|Not Authenticated|    ||            ||
+-----------------+    ||            ||
 || (7)   || (4)       ||            ||
 ||       \/           \/            ||
 ||     +----------------+           ||
 ||     | Authenticated  |<=++       ||
 ||     +----------------+  ||       ||
 ||       || (7)   || (5)   || (6)   ||
 ||       ||       \/       ||       ||
 ||       ||    +--------+  ||       ||
 ||       ||    |Selected|==++       ||
 ||       ||    +--------+           ||
 ||       ||       || (7)            ||
 \/       \/       \/                \/
+--------------------------------------+
|               Logout                 |
+--------------------------------------+
                  ||
                  \/
    +-------------------------------+
    |both sides close the connection|
    +-------------------------------+
Legend for the above diagram:

(1) connection without pre-authentication (OK 
greeting)
(2) pre-authenticated connection (PREAUTH 
greeting)
(3) rejected connection (BYE greeting)
(4) successful LOGIN or AUTHENTICATE command
(5) successful SELECT or EXAMINE command
(6) CLOSE or UNSELECT command, unsolicited 
CLOSED response code, or failed SELECT or 
EXAMINE command
(7) LOGOUT command, server shutdown, or 
connection closed
\end{verbatim}
\end{promptbox}
\caption{IMAP ASCII FSM}
\label{fig:imap-fsm}
\end{figure}

Fig. \ref{fig:imap-fsm} is the ASCII FSM of IMAP. Fig. \ref{fig:rtsp-init-table} is the Init State Transition Table of RTSP.

\begin{figure*}[t]
\centering
\begin{promptbox}
\begin{verbatim}
   +------------------+----------------+-----------+-------------------+
   | Action           | Requisite      | New State | Response          | 
   +------------------+----------------+-----------+-------------------+
   | SETUP            |                | Ready     | NRM=1, RP=0.0     |
   |                  |                |           |                   |
   | SETUP            | Needs Redirect | Init      | 3rr Redirect      |
   |                  |                |           |                   |
   | S -> C: REDIRECT | No Session hdr | Init      | Terminate all SES |
   +------------------+----------------+-----------+-------------------+
\end{verbatim}
\end{promptbox}
\caption{RTSP FSM table for state Init.}
\label{fig:rtsp-init-table}
\end{figure*}

Fig. \ref{fig:rtsp-init-table} is a state transition table for the Init state in RTSP, and LLMs need to understand it as: for the \texttt{Init} state, when condition (\texttt{Requisite}) happens, it should transition to \texttt{New State} and perform actions in \texttt{Action} as well as in \texttt{Response}.

\section{Prompt Details}
\label{app:prompt}

Task 1 question template prompt:

\begin{promptbox}
You are an expert in \var{protocol name} State Machine and you know the state machine of the protocol very well. I'll test your understanding of  \var{protocol name} State Machine. 

[optional: explain acronyms or state names if needed]

Assume your current state is \var{current\_state}, and you satisfy the conditions \var{condition}, can you predict which state you will transition to? Answer in the format <next\_state> next state </next\_state>.
Can you also predict if any resulting actions will happen, if there is nothing, answer <resulting\_action> NULL </resulting\_action>, otherwise, answer in the format <resulting\_action> action1, action2, ... </resulting\_action>.              
\end{promptbox}

Task 2 question template prompt looks like this:

\begin{promptbox}
You are an expert in \var{protocol name} State Machine and you know the state machine of the protocol very well. I'll test your understanding of  \var{protocol name} State Machine. 

[optional: explain acronyms or state names if needed]

Each state transition can be represented as a pair (previous\_state, condition, next\_state), meaning the system was in the previous\_state, the condition was satisfied, and it then moved to the next\_state.

Now I will give you the previous\_state \var{previous\_state}, and you transition to the next state \var{next\_state}, please infer all possible conditions that could have led to the transition. Only consider one-hop transitions — do not reason across multiple consecutive transitions.

Answer in the format: <condition> condition1, condition2 ... </condition>. 
\end{promptbox}

Task 3 question template prompt looks like this:

\begin{promptbox}
You are an expert in \var{protocol name} State Machine and you know the state machine of the protocol very well. I'll test your understanding of \var{protocol name}  State Machine. 

[optional: explain acronyms or state names if needed]

Each state transition can be represented as a pair (previous\_state, condition, next\_state), meaning the system was in the previous\_state, the condition was satisfied, and it then moved to the next\_state.

Now, I will give you the condition \var{condition} and the resulting state \var{next\_state}. Please infer all possible previous\_states that could have led to the transition. Only consider one-hop transitions — do not reason across multiple consecutive transitions.

Answer in the format: <previous\_state> state1, state2, ... </previous\_state>. 
    
\end{promptbox}

Task 4 question template prompt looks like this:
\begin{promptbox}
You are an expert in \var{protocol name} and you know the state machine of the protocol very well. I'll test your understanding of the protocol's State Machine.

[optional: explain acronyms or state names if needed]

Each state transition can be represented as a pair (previous\_state; condition; next\_state), meaning the system was in the previous\_state, the condition was satisfied, and it then moved to the next\_state.

Now I will give you the current\_state \var{current\_state} and a future\_state \var{future\_state}, please infer all possible paths by which the system could transition from the current\_state to the future\_state. 
Only consider simple paths, meaning no state may be visited more than once in a path. Do not include cycles, loops, or any transition sequence that revisits a previously visited state.

A simple path is a sequence:
current\_state; condition1; state1; condition2; state2; ...; future\_state

where:

- Each (state, condition, next\_state) corresponds to a single valid one-hop transition in the state machine.

- Do NOT collapse multiple transitions into one condition.

- Do NOT invent states that do not exist in the IMAP state machine.

- no state appears more than once,

- no cycles or loops are allowed,

- the path never returns to a previously visited state.

If there are multiple distinct possible paths, you MUST output all of them, each in its own <path>...</path> block.

Answer in the following format:

<path> current\_state; condition1; state1; condition2; state2; ...; future\_state </path>

<path> current\_state; condition1'; state1'; condition2'; state2'; ...; future\_state </path>
......

If there are no possible paths from current\_state to future\_state, please respond with:
<path> NULL </path>.

Below is an example, [the following part is per-protocol, e.g. for DHCP: ] 

if the current\_state is BOUND and the future\_state is INIT, and there are two possible paths from the current\_state to the future\_state, your answer should be:

<path> BOUND; timer T1 expires; RENEWING; timer T2 expires; REBINDING; receive DHCPNAK, Lease expired; INIT </path>

<path> BOUND; timer T1 expires; RENEWING; receive DHCPNAK; INIT </path>

\end{promptbox}

LLM judge prompt details for task 1, including separate prompts for judging next state equivalence and actions set equivalence (see Table \ref{tab:task-difficulty} for task1 metrics):
\begin{promptbox}
    You are an expert in \var{protocol name} and you know the state machine of the protocol very well. Can you decide if the below two states are equivalent?
The first state is the human generated ground truth next state: \var{ground truth next state}. The second state is the answer from a language model: \var{LLM answer next state}. 
Please answer in the format <equivalence>YES or NO</equivalence>. If you think they are not equivalent, please also provide the reason in the format <reason>your reason here</reason>.

You are an expert in \var{protocol name} and you know the state machine of the protocol very well. Can you decide if the below two sets of actions are equivalent?
The first set of actions is the human generated ground truth actions: \var{ground truth actions}. The second set of actions is the answer from a language model: \var{LLM answer actions}.
Please answer in the format <equivalence>YES or NO</equivalence>. If you think they are not equivalent, please also provide the reason in the format <reason>your reason here</reason>.

\end{promptbox}

LLM judge prompt details for task 2:

\begin{promptbox}
    You are an expert in \var{protocol name} and you know the state machine of the protocol very well. Can you decide if the below two sets of conditions are equivalent? 
The first set include the human generated ground truth conditions, and the second set include answers from a language model. The first set of conditions is: \var{ground truth condition(s)}. The second set of conditions is: \var{LLM answer condition(s)}. 
Plese answer in the format <equivalence>YES or NO</equivalence>. If you think they are not equivalent, please also provide the reason in the format <reason>your reason here</reason>. 
      
\end{promptbox}

LLM judge prompt details for task 3:
\begin{promptbox}
   You are an expert in \var{protocol name} and you know the state machine of the protocol very well. Can you decide if the below two sets of states are equivalent? 
The first set include the human generated ground truth states, and the second set include answers from a language model. The first set of states is: \var{ground truth previous state(s)}. The second set of states is: \var{LLM answer previous state(s)}. 
Plese answer in the format <equivalence>YES or NO</equivalence>. If you think they are not equivalent, please also provide the reason in the format <reason>your reason here</reason>. 
      
\end{promptbox}

LLM judge prompt details for task 4:

\begin{promptbox}
 You are an expert in \var{protocol name} and you know the state machine of the protocol very well. 
We are comparing two sets of state-machine paths.
Ground-truth paths: \var{ground truth path(s)}
LLM-generated paths: \var{LLM answer path(s)}
For each ground-truth path, determine whether it is exactly covered by the LLM-generated set.
For each LLM path, determine whether it matches at least one ground-truth path.
Two paths are considered equivalent only if their sequence of states and transition conditions are identical.

Answer in the format:
<covered\_gt\_count>NUMBER</covered\_gt\_count> 
<valid\_llm\_count>NUMBER</valid\_llm\_count>
\end{promptbox}

\section{Output Token Limit Effect}
\label{app:outlimit}
We examined the effect of output-length limits on reasoning performance. Using DeepSeek-R1 on Task 2, we varied the output token budget from 500 to 5,000 tokens, while our default setting uses 1,000 tokens. Under cosine-similarity evaluation, the score changed from 0.27 (500 tokens) to 0.31 (1,000 tokens) and 0.34 (5,000 tokens). Despite a 10 times increase in the output budget, the performance variation remained modest, suggesting that output-length limits are unlikely to be a major factor affecting our overall conclusions. Note that reasoning-token budgets are not exposed through the evaluated APIs and therefore cannot be controlled directly.

\section{LLM-as-a-Judge Reliability.} 
\label{app:llm-human}
Our judge performs semantic equivalence (paraphrase detection) rather than subjective quality assessment. Why cosine similarity is unreliable is that it often fails to recognize semantically equivalent protocol terminology. LLM judges compare the model's response (a short sequence-structured protocol information such as states or transitions) to a gold answer computed from the true model, making the task fundamentally different from conventional LLM-as-a-judge settings. To validate reliability, we randomly sampled 50 judgment instances and obtained independent annotations from three human annotators (we three authors), achieving 98\% agreement with the LLM judgments.

\section{Statistical Certainty: Bootstrap Results}
\label{app:stats}
We performed 1,000 protocol-level bootstrap resamples for all major task, model, and context comparisons and report 95\% confidence intervals together with paired bootstrap analyses. The bootstrap results distinguish robust findings from close numerical comparisons. For example, LLM-extracted-RFC and No-context partially overlap on Tasks 1–3. In contrast, larger gaps (e.g., Easy vs. Manual-RFC on Task 1: 0.945 [0.898, 0.982] vs. 0.516 [0.438, 0.589]) remain robust, supporting our main conclusions.

\section{Detailed Judge Score for Task 1, 2, 4}
Tables \ref{tab:task1-overall}, \ref{tab:task2-overall}, and \ref{tab:task4-overall} show the comparison of different judge scores for Tasks 1, 2, and 4.

\begin{table*}[ht]
\centering
\small
\caption{Judge score on Task~1 across answering LLMs.}
\label{tab:task1-overall}
\begin{tabular}{p{2cm}p{1cm}p{1cm}p{1cm}p{1cm}p{1cm}p{1cm}p{1cm}p{1cm}}
\toprule
Judge & gpt-4o-mini & qwen32B  & llama3.3-70B   & deepseek-r1 & grok-3-mini & gpt-5.2 & gemini-3-f-p & claude-s-4.5   \\
\midrule
Sim-0.6      &0.48 & 0.48  &  0.52      & 0.56&0.54 &0.55 & 0.58 & 0.59 \\
Llama3.3-70b&0.50  &0.52& 0.57 &0.60          &0.62 &0.65 & 0.64 & 0.66                    \\
 
\bottomrule
\end{tabular}
\end{table*}

\begin{table*}[ht]
\centering
\small
\caption{Judge score on Task~2 across answering LLMs.}
\label{tab:task2-overall}
\begin{tabular}{p{2cm}p{1cm}p{1cm}p{1cm}p{1cm}p{1cm}p{1cm}p{1cm}p{1cm}}
\toprule
Judge & gpt-4o-mini & qwen32B  & llama3.3-70B  & deepseek-r1 & grok-3-mini & gpt-5.2 & gemini-3-f-p & claude-s-4.5 \\
\midrule
Sim-0.6     &0.29 & 0.30  &  0.31  &0.31 &0.35 &0.29  & 0.41 & 0.40 \\
Llama3.3-70b&0.31 & 0.36  &  0.33  & 0.38 &0.44 &0.42  & 0.43 & 0.43\\

\bottomrule
\end{tabular}
\end{table*}

\begin{table*}[ht]
\centering
\small
\caption{Judge score on Task~4 across answering LLMs.}
\label{tab:task4-overall}
\begin{tabular}{p{2cm}p{1cm}p{1cm}p{1cm}p{1cm}p{1cm}p{1cm}p{1cm}p{1cm}}
\toprule
Judge &gpt-4o-mini & qwen32B  & llama3.3-70B  & deepseek-r1 & grok-3-mini & gpt-5.2 & gemini-3-f-p & claude-s-4.5  \\
\midrule
Llama3.3-70b &0.06 &0.33  &0.27  &0.38    &0.38 & 0.41 & 0.34    & 0.35 \\
Gemini-3-f-p &0.06 &0.33   &0.27    &0.40 &0.39 & 0.42  & 0.35    &0.38 \\

\bottomrule
\end{tabular}
\end{table*}

\section{Context Effect Results for Task 1, 3, 4}
Table \ref{tab:task1-context-effect}, \ref{tab:task3-context-effect}, and \ref{tab:task4-context-effect} shows the context effect results for tasks 1, 3, and 4.

\begin{table*}[ht]
\centering
\small
\caption{Effect of context type on Task~1 (smaller number is harder).}
\label{tab:task1-context-effect}
\begin{tabular}
{p{3cm}p{1cm}p{1cm}p{1cm}p{1cm}p{1cm}p{1cm}p{1cm}p{1cm}}
\toprule
Context Type & gpt-4o-mini & qwen32B  & llama3.3-70B & deepseek-r1 & grok-3-mini & gpt-5.2 & gemini-3-f-p & claude-s-4.5  \\
\midrule
Easy                 & 0.94  & 0.93  &0.95   &0.94 &0.95 &0.95 &0.95 &0.95 \\
Manual-extraced-RFC  & 0.43  & 0.41  &0.50   &0.55 &0.55 &0.56 &0.55 &0.59 \\
LLM-extracted-RFC    & 0.33  & 0.34  &0.39   &0.42 &0.42 &0.46 &0.49 &0.49 \\
No-context           &0.25   &0.30   &0.33   &0.41 &0.40 &0.43 &0.45 &0.48 \\
\bottomrule
\end{tabular}
\end{table*}

\begin{table*}[ht]
\centering
\small
\caption{Effect of context type on Task~3 (smaller number is harder).}
\label{tab:task3-context-effect}
\begin{tabular}
{p{3cm}p{1cm}p{1cm}p{1cm}p{1cm}p{1cm}p{1cm}p{1cm}p{1cm}}
\toprule
Context Type & gpt-4o-mini & qwen32B  & llama3.3-70B&deepseek-r1 & grok-3-mini & gpt-5.2 & gemini-3-f-p  &claude-s-4.5  \\
\midrule
Easy                &0.46 & 0.64  & 0.71 & 0.74  &0.86&0.86  & 0.86 &0.83 \\
Manual-extracted-RFC&0.14 & 0.30  & 0.32 &0.46  &0.45 &0.55 & 0.51 &0.52 \\
LLM-extracted-RFC   &0.10 & 0.21  & 0.28 &0.38  &0.38 &0.47 & 0.43 &0.40  \\
No-context          &0.09 & 0.14  & 0.20 & 0.36 &0.35 &0.41 & 0.36 &0.39 \\ 
\bottomrule
\end{tabular}
\end{table*}

\begin{table*}[ht]
\centering
\small
\caption{Effect of context type on Task~4 (smaller number is harder).}
\label{tab:task4-context-effect}
\begin{tabular}
{p{3cm}p{1cm}p{1cm}p{1cm}p{1cm}p{1cm}p{1cm}p{1cm}p{1cm}}
\toprule
Context Type & gpt-4o-mini& qwen32B  & llama3.3-70B & deepseek-r1 & grok-3-mini & gpt-5.2 & gemini-3-f-p & claude-s-4.5 \\
\midrule
Easy                 &0.14  &0.60   & 0.57  & 0.81&0.82 &0.78  & 0.70 &0.79 \\
Manual-extracted-RFC &0.03  &0.28  &0.19    & 0.27&0.25 &0.32  &0.26 &0.26   \\
LLM-extracted-RFC    &0.05  &0.25  &0.19    &0.23 &0.26 &0.29 & 0.24 &0.23  \\
No-context           &0.04  &0.19  &0.15    &0.25 &0.19 &0.26 & 0.19 & 0.19 \\
\bottomrule
\end{tabular}
\end{table*}

\section{Detailed Per-protocol Result for Task 1,2,3,4}
\begin{figure*}[htbp]
    \centering
    \includegraphics[width=\textwidth]{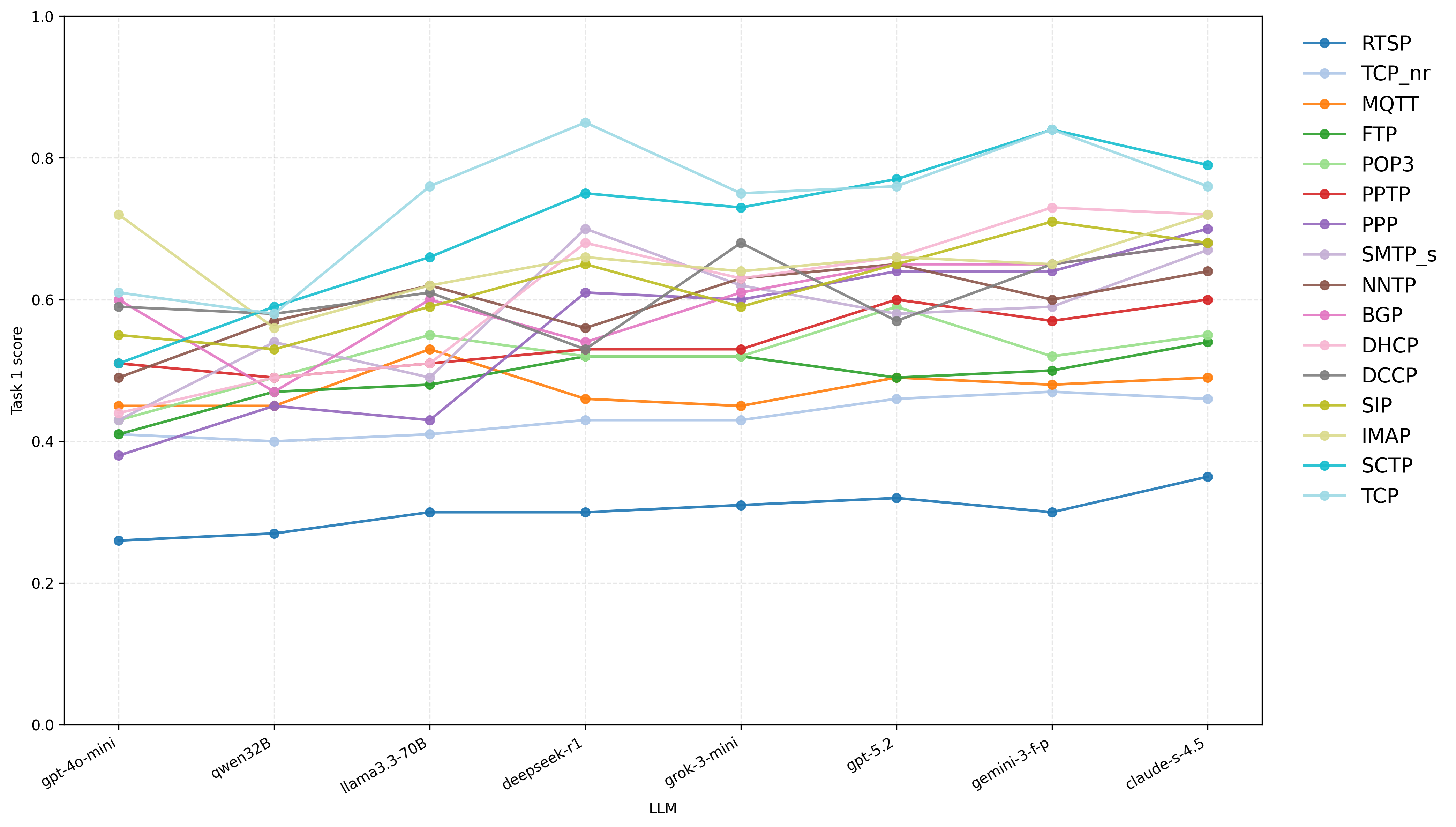}
    \caption{Protocol-level difficulty across LLMs (Task 1).}
    \label{fig:p1}
\end{figure*}

\begin{figure*}[htbp]
    \centering
    \includegraphics[width=\textwidth]{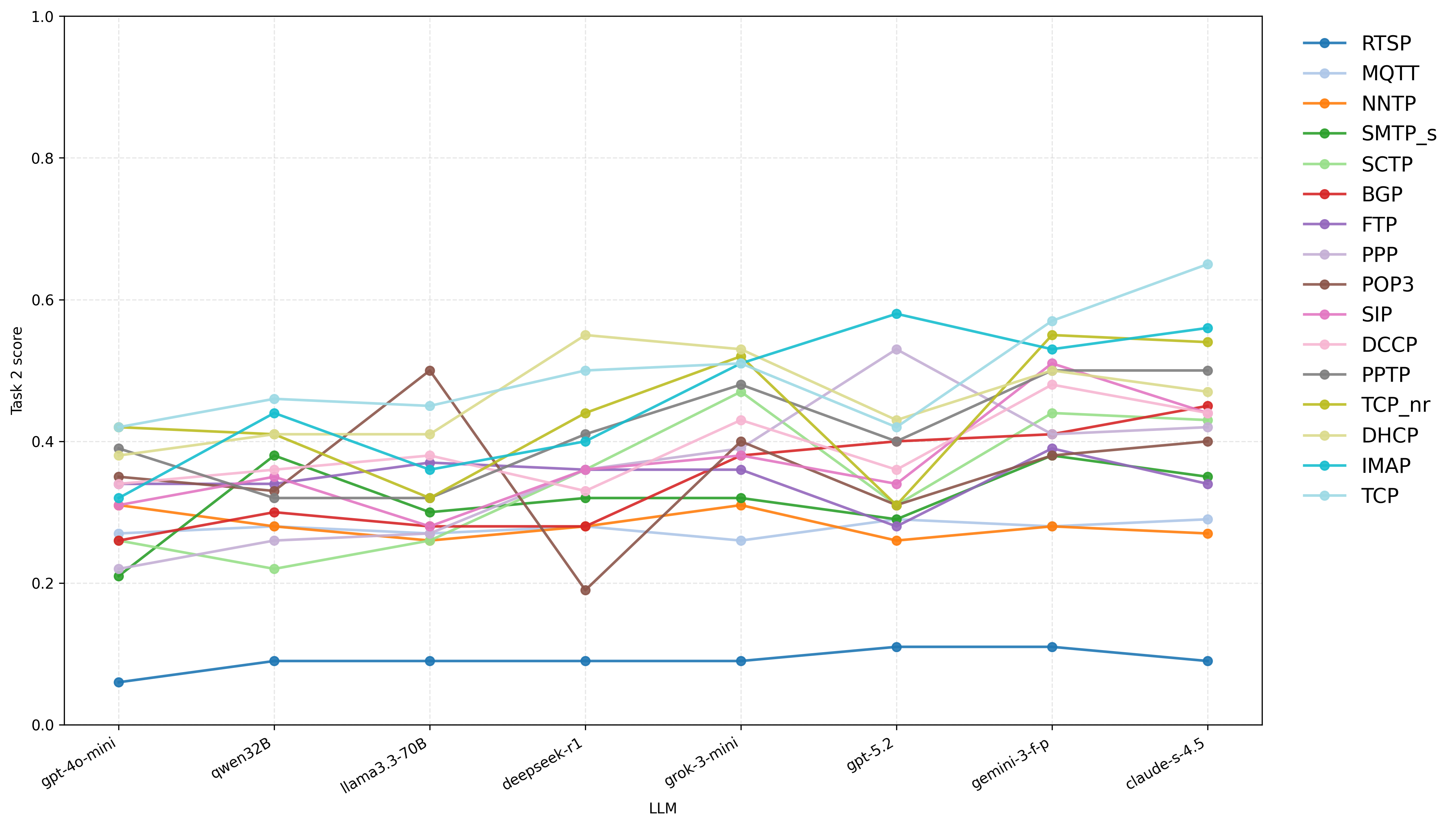}
    \caption{Protocol-level difficulty across LLMs (Task 2).}
    \label{fig:p2}
\end{figure*}

\begin{figure*}[htbp]
    \centering
    \includegraphics[width=\textwidth]{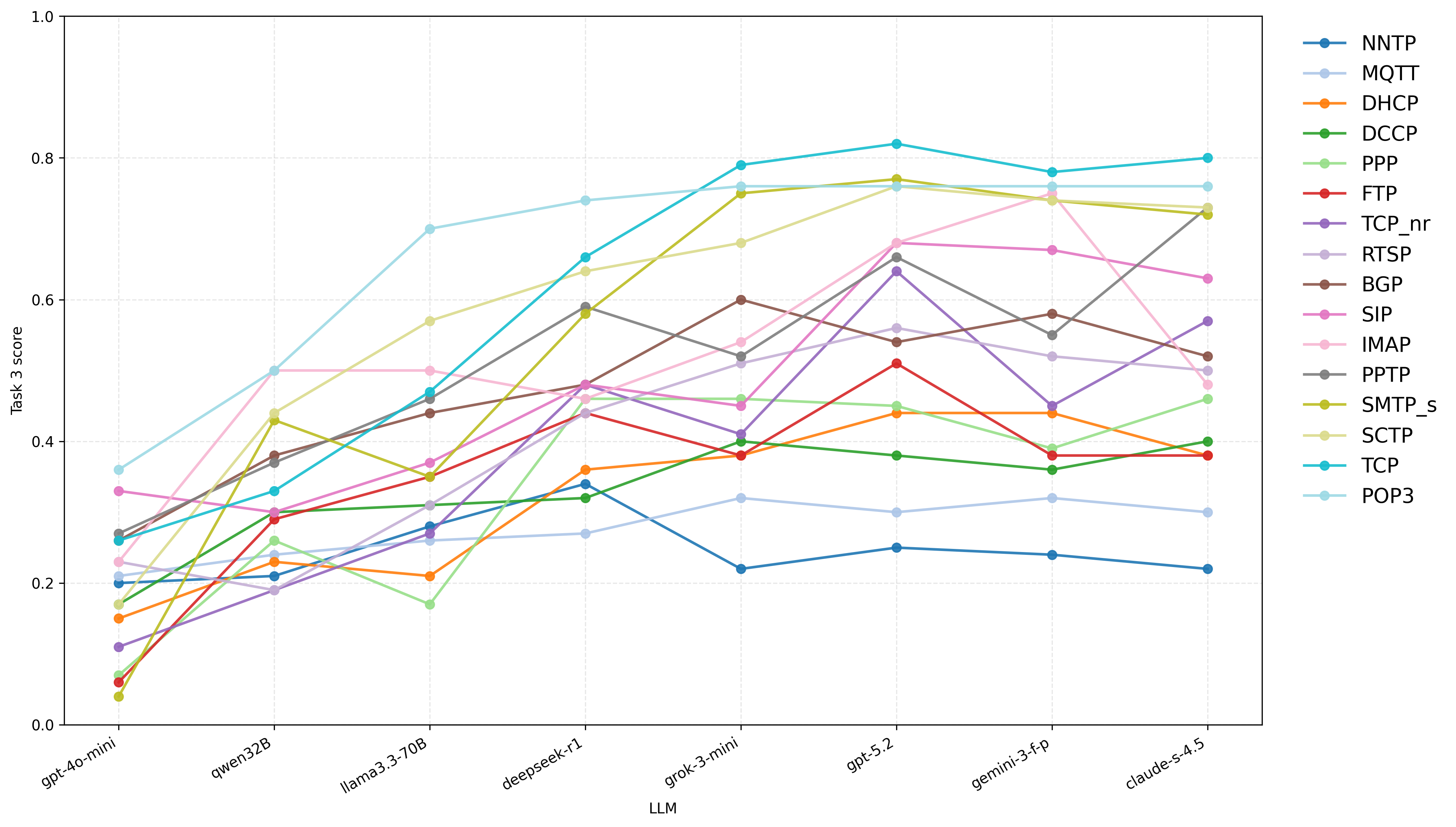}
    \caption{Protocol-level difficulty across LLMs (Task 3).}
    \label{fig:p3}
\end{figure*}

Table \ref{tab:task1-protocol-difficulty}, \ref{tab:task2-protocol-difficulty}, \ref{tab:task3-protocol-difficulty}, and \ref{tab:task4-protocol-difficulty} shows the detailed per-protocol results for Task 1,2,3,4. And Fig. \ref{fig:p1}, \ref{fig:p2}, \ref{fig:p3}, and \ref{fig:p4} showed the corresponding plots.

\begin{table*}[ht]
\centering
\small
\caption{Protocol-level difficulty on Task~1(ranked by Avg., smaller number is harder).}
\label{tab:task1-protocol-difficulty}
\begin{tabular}{p{1.5cm}p{1cm}p{1cm}p{1cm}p{1cm}p{1cm}p{1cm}p{1cm}p{1cm}p{0.5cm}p{0.5cm}}
\toprule
Protocol & gpt-4o-mini & qwen32B & llama3.3-70B  & deepseek-r1 & grok-3-mini & gpt-5.2 & gemini-3-f-p   & claude-s-4.5  & \textbf{Avg.} & \textbf{std.}\\
\midrule
RTSP    &0.26 &0.27 & 0.30   & 0.30 &0.31 &0.32  & 0.30 &0.35   &0.30  &0.03 \\
TCP\_nr &0.41 &0.40 &  0.41  & 0.43 &0.43 &0.46  & 0.47 & 0.46  &0.43 &0.02 \\
MQTT    &0.45 &0.45 & 0.53   &0.46  &0.45 &0.49  & 0.48 & 0.49  &0.48  &0.03 \\
FTP     &0.41 &0.47 & 0.48   & 0.52 &0.52 &0.49  & 0.50 &0.54   &0.49  &0.04\\
POP3    &0.43 &0.49 & 0.55   &0.52  &0.52 &0.59  & 0.52 & 0.55  &0.52  &0.04 \\
PPTP    &0.51 &0.49 & 0.51   &0.53  &0.53 &0.60  & 0.57 & 0.60  &0.54  &0.04 \\
PPP     &0.38 &0.45 & 0.43   &0.61  &0.60 &0.64  & 0.64 & 0.70  &0.56  &0.11 \\
SMTP\_s &0.43 &0.54 & 0.49    &0.70 &0.62 &0.58  & 0.59 & 0.67  & 0.58 &0.08 \\
NNTP    &0.49 &0.57 & 0.62   &0.56  &0.63 &0.65  & 0.60 & 0.64  &0.60 &0.05 \\
BGP     &0.60 &0.47 &  0.60  & 0.54 &0.61 &0.65  & 0.65 & 0.68  &0.60  &0.06 \\
DHCP    &0.44 &0.49 & 0.51    &0.68 &0.63 &0.66  & 0.73  &0.72  &0.61  &0.10 \\
DCCP    &0.59 &0.58 & 0.61   &0.53  &0.68 &0.57  & 0.65 & 0.68  &0.61  &0.05 \\
SIP     &0.55 &0.53 & 0.59   & 0.65 &0.59 &0.65  & 0.71 & 0.68  &0.62  &0.06 \\
IMAP    &0.72 &0.56 & 0.62   &0.66  &0.64 &0.66  & 0.65 & 0.72  &0.65  &0.05\\
SCTP    &0.51 &0.59 &  0.66  & 0.75 &0.73 &0.77  & 0.84 & 0.79  &0.70  &0.10 \\
TCP     &0.61 &0.58 & 0.76   &0.85  &0.75 &0.76  &0.84 & 0.76   &0.74  &0.09 \\
\bottomrule
\end{tabular}
\end{table*}

\begin{table*}[ht]
\centering
\small
\caption{Protocol-level difficulty on Task~2 (ranked by Avg., smaller number is harder).} 
\label{tab:task2-protocol-difficulty}
\begin{tabular}{p{1.5cm}p{1cm}p{1cm}p{1cm}p{1cm}p{1cm}p{1cm}p{1cm}p{1cm}p{0.5cm}p{0.5cm}}
\toprule
Protocol & gpt-4o-mini & qwen32B  & llama3.3-70B  & deepseek-r1& grok-3-mini & gpt-5.2  & gemini-3-f-p   & claude-s-4.5 & \textbf{Avg.} & \textbf{std.}\\
\midrule
RTSP     &0.06 & 0.09  & 0.09   &0.09&0.09 &0.11   &0.11 &0.09    &0.09  &0.01\\
MQTT     &0.27 & 0.28 & 0.27    &0.28&0.26 &0.29   & 0.28 &0.29   &0.28  &0.01 \\
NNTP     &0.31 & 0.28  & 0.26   &0.28&0.31 &0.26  & 0.28 &0.27    &0.28  &0.02 \\
SMTP\_s  &0.21 & 0.38 & 0.30    &0.32&0.32 &0.29   & 0.38 &0.35   &0.32  &0.05\\
SCTP     &0.26 & 0.22   & 0.26  &0.36&0.47 &0.31  & 0.44 &0.43    &0.34  &0.09\\
BGP      &0.26 & 0.30  & 0.28   &0.28&0.38 &0.40  & 0.41 &0.45    &0.34  &0.07 \\
FTP      &0.34 & 0.34  & 0.37   &0.36&0.36 &0.28   & 0.39 &0.34   &0.35  &0.03 \\
PPP      &0.22 & 0.26  & 0.27   &0.36&0.39 &0.53   &0.41  &0.42   & 0.36 &0.10 \\
POP3     &0.35 & 0.33  & 0.50   &0.19&0.40 &0.31   & 0.38 &0.40   &0.36  &0.08 \\
SIP      &0.31 & 0.35 &  0.28   &0.36&0.38 &0.34   & 0.51 &0.44   &0.37  &0.07\\
DCCP     &0.34 & 0.36 &  0.38   &0.33&0.43 &0.36   & 0.48 &0.44   & 0.39 &0.05 \\
PPTP     &0.39 & 0.32  & 0.32   &0.41&0.48 &0.40  & 0.50 &0.50    &0.42  &0.07 \\
TCP\_nr  &0.42 & 0.41 &  0.32   &0.44&0.52 &0.31  & 0.55&0.54     &0.44 &0.09 \\
DHCP     &0.38 & 0.41 & 0.41    &0.55&0.53 &0.43   & 0.50 &0.47   &0.46  &0.06 \\
IMAP     &0.32 & 0.44 & 0.36    &0.40&0.51 &0.58  & 0.53 &0.56    &0.46  &0.09\\
TCP      &0.42 & 0.46 &  0.45   &0.50&0.51 &0.42   & 0.57 &0.65   &0.50  &0.07\\

\bottomrule
\end{tabular}
\end{table*}

\begin{table*}[ht]
\centering
\small
\caption{Protocol-level difficulty on Task~3 (ranked by Avg., smaller number is harder).}
\label{tab:task3-protocol-difficulty}
\begin{tabular}{p{1.5cm}p{1cm}p{1cm}p{1cm}p{1cm}p{1cm}p{1cm}p{1cm}p{1cm}p{0.5cm}p{0.5cm}}
\toprule
Protocol & gpt-4o-mini  & qwen32B  & llama3.3-70B& deepseek-r1 & grok-3-mini & gpt-5.2 & gemini-3-f-p  & claude-s-4.5  & \textbf{Avg.} & \textbf{std.}\\
\midrule
NNTP    & 0.20 & 0.21  & 0.28   &0.34   &0.22 &0.25  & 0.24 & 0.22 &0.25  &0.04 \\
MQTT    & 0.21 & 0.24  &  0.26  &0.27   &0.32 &0.30  & 0.32 & 0.30 &0.28  &0.04 \\
DHCP    &0.15  & 0.23  & 0.21   &0.36   &0.38 &0.44  & 0.44  &0.38 &0.32  &0.10  \\
DCCP    & 0.17 & 0.30  &  0.31  &0.32   &0.40 &0.38  & 0.36 & 0.40 &0.33  &0.07 \\
PPP     & 0.07 & 0.26  & 0.17   &0.46   &0.46 &0.45  & 0.39 & 0.46 &0.34  &0.14  \\
FTP     & 0.06 & 0.29  & 0.35   &0.44   &0.38 &0.51  & 0.38 & 0.38 &0.35  &0.12  \\
TCP\_nr & 0.11 & 0.19  &  0.27  &0.48   &0.41 &0.64  & 0.45 & 0.57 &0.39  &0.17 \\
RTSP    & 0.23 & 0.19  &  0.31  &0.44   &0.51 &0.56  & 0.52& 0.50  &0.41  &0.13  \\
BGP     & 0.26 & 0.38  & 0.44   &0.48   &0.60 &0.54  & 0.58 & 0.52 &0.48  &0.11 \\
SIP     & 0.33 & 0.30  & 0.37   & 0.48  &0.45 &0.68  & 0.67 & 0.63 &0.49 &0.14 \\
IMAP    &0.23  & 0.50  & 0.50  &0.46    &0.54 &0.68  & 0.75 & 0.48 &0.52  &0.15 \\
PPTP    & 0.27 & 0.37  & 0.46   &0.59   &0.52 &0.66  & 0.55 & 0.73 &0.52  &0.14  \\
SMTP\_s & 0.04 & 0.43  & 0.35   &0.58   &0.75 &0.77  & 0.74 & 0.72 &0.55  &0.24 \\
SCTP    & 0.17 & 0.44  & 0.57   &0.64   &0.68 &0.76  & 0.74 & 0.73 &0.59  &0.19 \\
TCP     & 0.26 & 0.33  &  0.47  &0.66   &0.79 &0.82  & 0.78 & 0.80 &0.61  &0.21 \\
POP3    & 0.36 & 0.50  & 0.70   &0.74   &0.76 &0.76  & 0.76 & 0.76 &0.67  &0.14  \\
\bottomrule
\end{tabular}
\end{table*}

\begin{table*}[ht]
\centering
\small
\caption{Protocol-level difficulty on Task~4 (ranked by Avg., smaller number is harder).} 

\label{tab:task4-protocol-difficulty}
\begin{tabular}{p{1.5cm}p{1cm}p{1cm}p{1cm}p{1cm}p{1cm}p{1cm}p{1cm}p{1cm}p{0.5cm}p{0.5cm}}
\toprule
Protocol & gpt-4o-mini & qwen32B  & llama3.3-70B & deepseek-r1& grok-3-mini & gpt-5.2 & gemini-3-f-p   & claude-s-4.5  & \textbf{Avg.} & \textbf{std.}\\
\midrule
PPP    &0.02  &0.04  &0.06    &0.11   &0.10  &0.08  &0.11  &0.12    &0.08  &0.03      \\
BGP    &0.05  &0.08  &0.09    &0.13   &0.20  &0.14   &0.10  &0.15   &0.12  &0.04      \\
PPTP   &0.05  &0.07  &0.10    &0.20   &0.19  &0.16   &0.13  &0.18   &0.14  &0.05     \\
SCTP   &0.03  &0.07  &0.12    &0.19   &0.23  &0.19   &0.23  &0.20   &0.16  &0.07     \\
RTSP   &0.00   &0.21  &0.15    &0.25  &0.19  &0.21    &0.23  &0.23  &0.18  &0.08     \\
DCCP   &0.10   &0.17  &0.20   &0.26   &0.29  &0.23   &0.26  &0.25   &0.22  &0.06      \\
TCP\_nr&0.02  &0.22 &0.08  &0.31      &0.31  &0.34   & 0.24 &0.26   &0.22  &0.11      \\
MQTT   &0.06  &0.35  &0.21    &0.25   &0.27  &0.52   &0.21  &0.26   &0.27  &0.12      \\
FTP    &0.06  &0.38  &0.27    &0.26   &0.27  &0.52   &0.24  &0.26   &0.28  &0.12      \\
DHCP   &0.03  &0.34  & 0.27   &0.43  &0.43  &0.43    & 0.25 &0.33  &0.31 &0.13      \\
NNTP   &0.10   &0.57  &0.39    &0.43  &0.52  &0.49   &0.33  &0.46   &0.41  &0.14    \\
TCP    &0.04  &0.47  &0.36    &0.53   &0.40  &0.62   & 0.40  &0.42  &0.41  &0.16      \\
SMTP\_s&0.06  &0.47  &0.30    &0.52   &0.52  &0.46   & 0.58  &0.48  &0.42  &0.16     \\
SIP    &0.08  &0.51  &0.54    &0.68   &0.64  &0.65   &0.65  &0.68   &0.55  &0.19      \\
POP3   &0.12 &0.58  &0.52    &0.79   &0.71  &0.75   & 0.73  &0.77  &0.62  &0.21      \\
IMAP   &0.21   &0.71  &0.69   &0.89   &0.86  &0.82  & 0.82  &0.84   &0.73  &0.21     \\
\bottomrule
\end{tabular}
\end{table*}

Table \ref{tab:overall-protocol-difficulty} shows per-protocol performance for each task, and the average score and standard deviation value across tasks for each protocol.

\begin{table*}[t]
\centering
\small
\caption{Overall protocol score across tasks. Smaller values indicate harder protocols. Avg. is the average score over Tasks 1, 2, 3, and 4, and Std. measures difficulty variability across tasks.} 
\label{tab:overall-protocol-difficulty}
\begin{tabular}{lccccccc}
\toprule
Rank & Protocol & Task~1 & Task~2 & Task~3 & Task~4 & Avg. & Std. \\
\midrule
1  & RTSP     & 0.30 & 0.09 & 0.41 &0.18   &0.25  &0.12  \\
2  & MQTT     & 0.48 & 0.28 & 0.28 &0.27   &0.33  &0.08  \\
3  & PPP      & 0.56 & 0.36 & 0.34 &0.08  &0.34  &0.17  \\
4  & TCP\_nr & 0.43 & 0.44 & 0.39  &0.22   &0.37  &0.09  \\
5 & FTP      & 0.49 & 0.35 & 0.35  &0.28   &0.37  &0.08 \\
6 & BGP      & 0.60 & 0.34 & 0.48  &0.12    &0.39  &0.18 \\
7 & DCCP     & 0.61 & 0.39 & 0.33  &0.22  &0.39  &0.14  \\
8  & NNTP     & 0.60 & 0.28 & 0.25 &0.41  &0.39  &0.14  \\
9 & PPTP     & 0.54 & 0.42 & 0.52  &0.14    &0.41  &0.16 \\
10 & DHCP     & 0.61 & 0.46 & 0.32 &0.31   &0.43  &0.12  \\
11 & SCTP     & 0.70 & 0.34 & 0.59 &0.16   &0.45  &0.21  \\
12 & SMTP\_s & 0.58 & 0.32 & 0.55  &0.42   &0.47  &0.11  \\
13  & SIP      & 0.62 & 0.37 &0.49 &0.55   &0.51  &0.09  \\ 
14 & POP3     & 0.52 & 0.36 & 0.67 &0.62   &0.54  &0.11  \\
15 & TCP      & 0.74 & 0.50 & 0.61 &0.41   &0.57 & 0.12 \\
16 & IMAP     & 0.65 & 0.46 & 0.52 &0.73   &0.59  &0.10  \\

\bottomrule
\end{tabular}
\end{table*}

\section{Detailed Results Task 4}

 Table \ref{tab:task4-protocol-prf} displays the Precision, Recall, and F1 results of Task~4 across LLMs and protocols.

\begin{table*}[ht]
\centering
\scriptsize
\caption{Protocol-level Precision, Recall, F1 on Task~4: P/R/F1 represents Precision/Recall/F1. The ranking is the same as in Table \ref{tab:task4-protocol-difficulty}, i.e., ranked by avg. score in task 4 from low to high. We exclude the column for gpt-4o-mini to save space because its performance on task 4 is very low.} 

\label{tab:task4-protocol-prf}
\begin{tabular}{p{0.9cm}p{1.7cm}p{1.7cm}p{1.6cm}p{1.6cm}p{1.6cm}p{1.6cm}p{1.6cm}}
\toprule
Protocol & qwen32B (P/R/F1)  & llama3.3-70B (P/R/F1) & deepseek-r1 (P/R/F1) & grok-3-mini(P/R/F1)& gpt-5.2 (P/R/F1) & gemini-3-f-p (P/R/F1)   & claude-s-4.5 (P/R/F1)  \\
\midrule
PPP       &0.10/0.00/0.01    &0.17/0.02/0.03   &0.26/0.05/0.09 &0.22/0.04/0.07 &0.18/0.01/0.02    &0.26/0.05/0.08  &0.32/0.04/0.07    \\
BGP       &0.29/0.03/0.06    &0.25/0.06/0.10   &0.29/0.11/0.15 &0.21/0.10/0.13 &0.33/0.06/0.10     &0.25/0.11/0.15 &0.26/0.08/0.12   \\
PPTP      & 0.19/0.05/0.08   &0.28/0.14/0.18   &0.29/0.22/0.24 &0.31/0.22/0.25 &0.28/0.16/0.20    &0.29/0.18/0.22  &0.30/0.23/0.26    \\
SCTP      &0.27/0.04/0.08    &0.45/0.10/0.16   &0.50/0.18/0.25 &0.57/0.24/0.32 &0.46/0.13/0.20    &0.57/0.21/0.30  &0.54/0.23/0.32   \\
RTSP      &0.31/0.11/0.16    &0.31/0.15/0.20   &0.21/0.22/0.21 &0.19/0.19/0.19 &0.27/0.13/0.17   & 0.27/0.24/0.26  &0.29/0.31/0.30     \\
DCCP      &0.32/0.09/0.14    &0.35/0.14/0.20   &0.41/0.24/0.30 &0.44/0.26/0.32 &0.40/0.18/0.24    &0.48/0.25/0.32  &0.40/0.26/0.31    \\
TCP\_cgc  &0.35/0.25/0.29    &0.31/0.25/0.28   &0.44/0.41/0.42 &0.44/0.42/0.43 &0.46/0.40/0.42    &0.36/0.36/0.36  &0.39/0.37/0.38    \\
MQTT      &0.33/0.30/0.31    &0.22/0.24/0.23   &0.26/0.26/0.26 &0.27/0.27/0.27 &0.49/0.46/0.48    &0.21/0.26/0.23  &0.26/0.27/0.26     \\
FTP       &0.39/0.20/0.27    &0.28/0.23/0.25   &0.23/0.24/0.23 &0.28/0.28/0.28 &0.50/0.32/0.38    &0.24/0.29/0.26  &0.28/0.28/0.27     \\
DHCP      &0.43/0.30/0.35    &0.38/0.33/0.35   &0.47/0.45/0.46 &0.43/0.45/0.44 &0.45/0.39/0.41    &0.36/0.35/0.35  &0.38/0.41/0.39    \\
NNTP      &0.55/0.51/0.52    &0.36/0.44/0.39   &0.43/0.49/0.45 &0.47/0.55/0.50 &0.48/0.44/0.46   &0.28/0.39/0.31   &0.36/0.51/0.40    \\
TCP       &0.50/0.35/0.41    &0.38/0.36/0.37   &0.47/0.48/0.47 &0.39/0.41/0.40 &0.61/0.56/0.58    & 0.35/0.54/0.42 &0.41/0.45/0.43     \\
SMTP\_s   &0.51/0.34/0.40    &0.41/0.27/0.33   &0.57/0.48/0.52 &0.50/0.49/0.49 &0.55/0.37/0.43   &0.71/0.55/0.62   &0.54/0.56/0.55    \\
SIP       &0.59/0.26/0.36    &0.59/0.37/0.45   &0.68/0.49/0.56 &0.72/0.45/0.54 &0.64/0.41/0.50    &0.66/0.52/0.58  &0.60/0.43/0.50    \\
POP3      &0.60/0.56/0.58    &0.48/0.54/0.50   &0.79/0.75/0.77 &0.68/0.64/0.66 &0.88/0.81/0.83   &0.79/0.76/0.77   &0.82/0.71/0.75     \\
IMAP      &0.77/0.60/0.67    &0.67/0.69/0.68   &0.88/0.89/0.89 &0.80/0.78/0.79 &0.82/0.75/0.78    &0.75/0.77/0.76  &0.81/0.81/0.81   \\
\bottomrule
\end{tabular}
\end{table*}


\label{app:t4gemini}

Table \ref{tab:task4-gemini3fp} shows the detailed results of Gemini-3-flash-preview answering Task 4, specifically where we explored several cosine similarity thresholds used as judges, as well as the two LLM judges. We observed that no matter how we tune the cosine similarity thresholds, the judge scores still show a huge gap with the two LLM judges. 

\begin{table*}[ht]
\centering
\scriptsize
\renewcommand{\arraystretch}{0.9}
\caption{Task 4 results (Gemini-3-flash-preview). } 

\begin{tabular}{p{2cm}|p{3cm}|p{2cm}|p{3.3cm}|p{3cm}}
\toprule
Protocol & Context type & \#Queries & \#Correct paths set (perfect match) \newline 
(judges: 0.6-sim, 0.7-sim, 0.8-sim, llama3.3-70b , gemini3fp) & Micro Precision, Micro Recall, F1 (judge: gemini3fp)\\
\midrule
TCP & Easy table \newline Manually-extracted rfc \newline  LLM-extracted \newline No context
    & 110
     & 86, 86, 86, 66, 67  \newline 67, 66, 59, 45, 42 \newline 65, 62, 58, 39, 38 \newline 60, 58, 56, 36, 31 & 0.61, 0.81, 0.69 \newline 0.31, 0.54, 0.40 \newline 0.26, 0. 52, 0.35 \newline 0.26, 0.43, 0.32 \\
\midrule
TCP new reno & Easy table \newline Manually-extracted rfc  \newline LLM-extracted \newline No context
    & 12
    & 12, 12, 12, 11, 11 \newline 11, 11, 7, 2, 0\newline 10, 9, 4, 2, 0  \newline 11, 10, 8, 4, 0 & 0.93, 1.0, 0.96 \newline 0.16, 0.15, 0.16 \newline 0.10, 0.08, 0.09 \newline 0.14, 0.12, 0.12  \\
\midrule
SCTP & Easy table \newline Manually-extracted rfc  \newline LLM-extracted \newline No context
    & 56
     & 47, 46, 42, 26, 25\newline 44, 40, 33, 17, 11 \newline 44, 40, 35, 10, 7 \newline 45, 43, 37, 8, 8 & 0.73, 0.39, 0.51 \newline 0.79, 0.26, 0.39 \newline 0.60, 0.14, 0.23 \newline 0.50, 0.15, 0.23 \\
\midrule
DCCP & Easy table \newline Manually-extracted rfc \newline LLM-extracted \newline No context
    & 72
    & 64, 60, 53, 36, 37 \newline 58, 45, 35, 23, 19 \newline 53, 36, 23, 25, 13  \newline 60, 53, 42, 10, 10 & 0.92, 0.50, 0.70 \newline 0.56, 0.28, 0.37 \newline 0.48, 0.21, 0.29 \newline 0.24, 0.14, 0.18 \\
\midrule
PPP & Easy table \newline Manually-extracted rfc \newline LLM-extracted \newline No context
    & 90
    & 86, 77, 62, 20, 21 \newline 64, 39, 8, 11, 10  \newline 46, 18, 4, 7, 5 \newline 77, 54, 27, 8, 6 & 0.89, 0.17, 0.29 \newline 0.31, 0.07, 0.11 \newline 0.29, 0.06, 0.10 \newline 0.19, 0.03, 0.06 \\
\midrule
PPTP & Easy table \newline Manually-extracted rfc \newline LLM-extracted \newline No context
    & 72
    & 72, 72, 64, 34, 38 \newline 67, 55, 28, 4, 1 \newline 41, 31, 17, 3, 1 \newline 71, 67, 41, 3, 1 & 0.96, 0.62, 0.75 \newline 0.12, 0.07, 0.09 \newline 0.13, 0.04, 0.07 \newline 0.09, 0.07, 0.08\\
\midrule
DHCP & Easy table \newline Manually-extracted rfc \newline LLM-extracted \newline No context
    & 56
    & 52, 49, 44, 34, 34\newline 43, 35, 19, 25, 8  \newline 41, 31, 15, 24, 9 \newline 33, 30, 22, 14, 6 & 0.79, 0.70, 0.74 \newline 0.30, 0.28, 0.29 \newline 0.33, 0.33, 0.33 \newline 0.13, 0.19, 0.15  \\
\midrule
BGP & Easy table \newline Manually-extracted rfc \newline LLM-extracted \newline No context
    & 30
    & 27, 25, 23, 6, 7\newline 24, 18, 9, 4, 1\newline 23, 14, 4, 3, 0 \newline 23, 12, 5, 1, 1 & 0.80, 0.33, 0.47 \newline 0.29, 0.17, 0.21 \newline 0.06, 0.02, 0.02 \newline 0.04, 0.02, 0.02\\
\midrule
SMTP server & Easy table \newline Manually-extracted rfc \newline LLM-extracted \newline No context
    & 30
    & 30, 30, 30, 27, 27 \newline 27, 25, 21, 17, 16 \newline 27, 25, 22, 15, 14  \newline 29, 28, 23, 18, 12 & 1.0, 0.83, 0.91 \newline 0.68, 0.58, 0.62 \newline 0.63, 0.5, 0.56 \newline 0.51, 0.35, 0.41\\
\midrule
POP3 & Easy table \newline Manually-extracted rfc \newline LLM-extracted \newline No context
    & 6
    & 6, 6, 6, 6, 6 \newline 6, 5, 4, 4, 4\newline 6, 5, 4, 5, 4 \newline 5, 5, 4, 4, 3 & 1.0, 1.0, 1.0 \newline 0.8, 0.89, 0.84 \newline 0.71, 0.56, 0.63 \newline 0.67, 0.67, 0.67 \\
\midrule
IMAP & Easy table \newline Manually-extracted rfc \newline LLM-extracted \newline No context
    & 20
    & 20, 20, 20, 20, 20 \newline  20, 20, 20, 19, 17 \newline 20, 20, 20, 20, 16\newline 18, 18, 17, 15, 13 & 1.0, 1.0, 1.0 \newline 0.87, 0.87, 0.87 \newline 0.84, 0.84, 0.84 \newline 0.43, 0.48, 0.45 \\
\midrule
NNTP & Easy table \newline Manually-extracted rfc \newline LLM-extracted \newline No context
    & 72
    & 68, 68, 68, 65, 65  \newline 42, 36, 22, 12, 10 \newline 44, 40, 24, 14, 12 \newline 42, 38, 23, 10, 7 & 0.86, 0.95, 0.90 \newline 0.07, 0.18, 0.10 \newline 0.08, 0.17, 0.11 \newline 0.07, 0.23, 0.11\\
\midrule
SIP & Easy table \newline Manually-extracted rfc \newline LLM-extracted \newline No context
    & 20
    & 20, 20, 20, 18, 20 \newline 19, 19, 14, 14, 11 \newline 19, 14, 10, 12, 10  \newline 18, 15, 14, 12, 10 & 1.0, 1.0, 1.0 \newline 0.70, 0.55, 0.62 \newline 0.69, 0.38, 0.49 \newline 0.44, 0.30, 0.36\\
\midrule
RTSP & Easy table \newline Manually-extracted rfc \newline LLM-extracted \newline No context
    & 6
    & 6, 5, 5, 3, 5  \newline 5, 3, 2, 2, 2 \newline 2, 0, 0, 1, 0 \newline 1, 1, 0, 0, 0 & 1.0, 0.92, 0.96 \newline 0.64, 0.56, 0.60 \newline 0.0, 0.0, 0 \newline 0.0, 0.0, 0\\
\midrule
FTP & Easy table \newline Manually-extracted rfc \newline LLM-extracted \newline No context
    & 90
    & 85, 82, 79, 74, 75 \newline 51, 47, 39, 4, 2 \newline 55, 52, 40, 12, 10  \newline 50, 47, 37, 2, 2 & 0.84, 0.95, 0.89 \newline 0.05, 0.10, 0.07 \newline 0.08, 0.10, 0.09 \newline 0.06, 0.12, 0.08\\
\midrule
MQTT & Easy table \newline Manually-extracted rfc \newline LLM-extracted \newline No context
    & 132
    & 116, 115, 114, 104, 109  \newline 74, 58, 28, 9, 2 \newline 72, 63, 38, 12, 0 \newline 72, 57, 31, 11, 1 & 0.78, 0.94, 0.85 \newline 0.02, 0.05, 0.02 \newline 0.01, 0.02, 0.01 \newline 0.01, 0.03, 0.02\\

\bottomrule
\end{tabular}
\label{tab:task4-gemini3fp}
\end{table*}

\section{Formulas of R/P/F1 in task 4}
\label{app:formulas}
Full details of the formulas to compute $Recall_{p,c}$, $Precision_{p,c}$, and $F1_{p,c}$ are given below:

\begin{equation}
  Recall_{p,c}
=
\frac{total\_recall\_hits_{p,c}}
{total\_GT_p}
\end{equation}
\begin{equation}
Precision_{p,c}
=
\frac{total\_precision\_hits_{p,c}}
{total\_Pred_{p,c}}
\end{equation}
\begin{equation}
F1_{p,c}
=
\frac{2 \cdot Precision_{p,c} \cdot Recall_{p,c}}
{Precision_{p,c} + Recall_{p,c}}
\end{equation}

\end{document}